\documentclass{article}

\usepackage{benstyle,enumitem}

\usepackage[main, final]{neurips_2026}

\usepackage[utf8]{inputenc} %
\usepackage[T1]{fontenc}    %
\usepackage{hyperref}       %
\usepackage{url}            %
\usepackage{booktabs}       %
\usepackage{amsfonts}       %
\usepackage{nicefrac}       %
\usepackage{microtype}      %
\usepackage{xcolor}         %
\usepackage{amsmath}
\usepackage{amssymb}
\usepackage{xcolor}
\usepackage{algorithm,algpseudocode}
\usepackage{wrapfig}
\usepackage{lipsum}

\newcommand{\sfext}[0]{\mathsf{ext}}

\title{Christoffel-DPS: Optimal sensor placement in diffusion posterior sampling for arbitrary distributions}

\author{%
James Rowbottom\footnotemark[1] \\ Department of Applied Mathematics and Theoretical Physics \\ University of Cambridge \\ \texttt{jr908@cam.ac.uk}
\And
Nick Huang\footnotemark[1] \\ Department of Mathematics \\ Simon Fraser University\\ \texttt{nick\_huang@sfu.ca}
\And
Carola-Bibiane Sch\"onlieb \\ Department of Applied Mathematics and Theoretical Physics \\ University of Cambridge \\ \texttt{cbs31@cam.ac.uk}
\And
Ben Adcock \\ Department of Mathematics \\ Simon Fraser University \\ \texttt{ben\_adcock@sfu.ca}
}

\begin{document}

\maketitle
\def\thefootnote{*}\footnotetext{Equal contribution}\def\thefootnote{\arabic{footnote}}

\begin{abstract}
State estimation is a critical task in scientific, engineering and control applications. Since the reliability of reconschritructions can depend on the number and position of sensors, Optimal sensor placement (OSP) is essential in scenarios where measurements are sparse and expensive. However, classical OSP approaches rely on Gaussian assumptions and are consequently unable to account for complex distributions encountered in many real-world systems. Generative-model-based reconstruction using sensor guided diffusion posterior sampling (DPS) has emerged as a promising technique for reconstructing states from highly complex distributions. Existing approaches to sensor selection either choose an unrealistically large number of sensors or employ strategies that emulate classical OSP methods. This results in a mismatch, wherein new models are paired with classical OSP tools, and motivates the need for fundamentally new ideas towards OSP that match the recent advances made in powerful recovery models. In this work, we introduce a distribution-free sensor placement framework based on the Christoffel function. Our main theoretical contributions introduce a mathematical formulation of optimal sampling and recovery guarantees for posterior sampling with arbitrary sensors and signal distributions. We use these to derive a new OSP strategy with non-asymptotic bounds on the number of sensors needed for recovery. The Christoffel function provides the mathematical foundation to maximise the identifiability of states by exploiting the geometry of the state distribution. This yields optimal sensor locations that parameterise a posterior that minimises the $L^2$ reconstruction error of the state estimate over the prior distribution of states. Building on this, we develop \textbf{Christoffel-DPS}, with both offline and online variants, that implements nonparametric realizations of Christoffel sampling for generative models. As we show, Christoffel-DPS outperforms Gaussian OSP baselines and existing generative-model-based placement methods, validating that distribution-free sensing is both theoretically principled and practically superior. The framework is model agnostic, and we demonstrate its application to a range of unconditional DPS and flow matching models on structurally non-Gaussian benchmarks, showing the efficacy of Christoffel-DPS in low sensor budget regimes.
\end{abstract}

\section{Introduction}

State reconstruction from sparse measurements is an important problem across science and engineering. Applications include
ocean state estimation from networks of buoys \citep{yildirim2009ocean}, atmospheric data assimilation from weather stations \citep{majumdar2016targeted}, fluid-flow reconstruction around aerodynamic surfaces from pressure taps \citep{willcox_unsteady_2006}, seismic imaging from receiver arrays \citep{maurer2010survey} and biomedical imaging from compressive measurements \citep{soler2022eeg}. In such settings, sensors are expensive to deploy and operate, the signal lives in a high- or infinite-dimensional state space, and the number and location of sensors can be the dominant factors controlling reconstruction accuracy. Optimal sensor placement (OSP) asks how to choose these measurements so as to maximize reconstruction accuracy under a prescribed sensing budget.

Classical approaches to OSP are closely tied to linear recovery
methods, reduced-order modelling and Gaussian assumptions. In Bayesian
optimal experimental design (OED), standard A-, D- and E-optimality
criteria optimize functionals of the posterior covariance matrix.
Typical methods, including gappy POD
\citep{everson_karhunenloeve_1995,willcox_unsteady_2006}
Gaussian-process and kriging regression \citep{krause_near-optimal_2008},
ensemble Kalman filtering, and reduced-order POD-Galerkin, DEIM/QDEIM
models \citep{chaturantabut_nonlinear_2010,drmac_new_2016} and SSPOR
\citep{manohar_data-driven_2018}, choose sensors so that a reduced
basis is well conditioned when restricted to the sensor measurements.
These methods have been highly successful and often come with
algorithmic and theoretical guarantees. However, their placement
criteria are typically derived from linearity and Gaussian
assumptions, and as a result they do not directly exploit the
geometry of complex signal distributions such as multi-modal,
strongly non-Gaussian or manifold-supported priors.

This limitation has become increasingly important with the rapid growth of
learned recovery operators. Supervised inverse maps from sparse measurements to
full fields are the natural starting point: Voronoi-tessellation CNNs
\citep{fukami_global_2021}, transformer reconstructors such as the
Senseiver \citep{santos_development_2023} and the Energy Transformer
\citep{zhang_operator_2025}, DeepONet-style operator networks
\citep{dang_deep_2025}, and graph-transformer reconstructors on
unstructured meshes \citep{duthe_graph_2025-2}. Generative
reconstructors take a probabilistic view: PhySense
\citep{ma_physense_2025} trains a conditional flow-matching
reconstructor on randomised sensor layouts and optimizes placement
post-hoc by projected gradient descent against a reconstruction loss;
SDIFT \citep{chen_generating_2025} runs sequential diffusion in a
learned functional-Tucker latent with message-passing posterior
sampling for irregular sparse observations; DiffusionPDE
\citep{huang_diffusionpde_2024} and FunDPS keep an unconditional
diffusion prior and add measurement and PDE-residual guidance at
sampling time, with ConFIG \citep{amoros-trepat_guiding_2026}
introducing conflict-free gradient projections to stabilise
multi-objective guidance, DDO \citep{lim_score-based_2023} working
directly in function space via Cameron--Martin geometry, and DDIS
\citep{lin2026decoupled} decoupling joint-state training by
enforcing the PDE constraint with a separately learned operator at
sampling time.

Overall, these models are far from the setting of classical OSP. Generative
models, and in particular diffusion models trained on full-state
samples \citep{karras_elucidating_2022}, encode highly expressive
non-Gaussian priors and admit recovery through techniques such as
diffusion posterior sampling (DPS) \citep{chung2023diffusion} and
various others \citep{baldassari2026preconditioned}. In the limit of a
perfectly trained denoiser the prior acts, for our purposes, as an
oracle, concentrating probability mass on the manifold of plausible
states and rendering the posterior highly non-Gaussian. Despite this
shift, OSP for generative recovery operators is comparatively
under-developed. PhySense's projected-gradient stage
\citep{ma_physense_2025} is the only fully end-to-end placement loop,
but its reconstruction-loss objective is mean-squared and its sensor
optimisation reverts to a Gaussian-likelihood treatment under its
modelling assumptions, rendering it essentially equivalent to
A-optimal design. Ensemble-based approaches such as the per-pixel
standard-deviation score of \citet{chakraborty_adaptive_2026}, the
gradient-weighted class-activation map of \citet{xu_optimal_2024},
the cartoonist-style uncertainty of \citet{karczewski_diffusion_2024},
and the EnKF-driven neural network of \citet{deng_deep_2021} are also
A-optimal in spirit, scoring sensors by a second-moment summary of an
ensemble; attention-based placement \citep{zhao_optimal_2025} and the
curriculum reformulation of \citet{marcato_journey_2024} fall in the
same broadly Gaussian family. None are distribution-free in the
formal sense, and none come with non-asymptotic recovery guarantees
that match the expressivity of the underlying generative prior.

\begin{wrapfigure}{r}{0.5\linewidth}
  \centering
  \includegraphics[width=\linewidth]{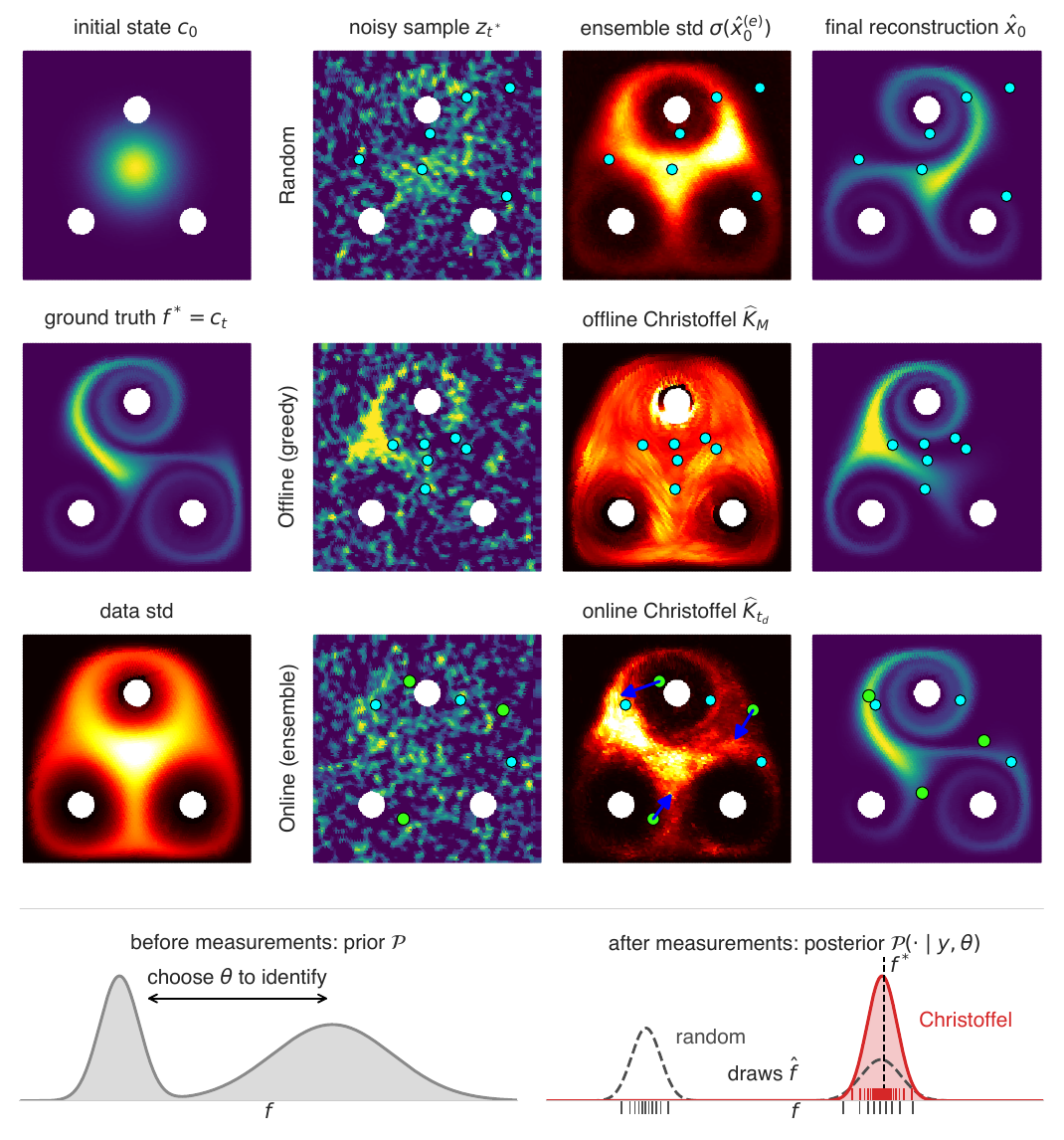}
  \caption{\textbf{Offline and online Christoffel-DPS.} Three DPS sampling trajectories on Pinball ~\S\ref{ss:pinball}. \textbf{Row 1 (random):} 6 fixed sensors at random positions. \textbf{Row 2 (offline):} 6 fixed sensors ($\textcolor{cyan}{\bullet}$) from greedy Christoffel-DPS, one DPS draw. \textbf{Row 3 (online):} 3 fixed anchors ($\textcolor{cyan}{\bullet}$) located via sampling offline Christoffel function, 3 mobile sensors ($\textcolor{green}{\bullet}$) chosen online by the ensemble-greedy update. \textbf{Columns}: initial state, ground truth, data std; noisy state $z_{t^*}$ (illustrative); row 1 pixel std of the Tweedie estimates, $\sigma(\hat{x}_0)$, row 2 offline empirical Christoffel score, row 3 online interim score driving mobile sensor drift; final reconstruction at $T=0$. \textbf{Bottom (schematic):} prior and posterior when sensors do (Christoffel) or do not (random) identify the correct mode.}
  \label{fig:F1}
\end{wrapfigure}

This paper aims to close this gap. We introduce a
theoretical framework for OSP which provides non-asymptotic recovery
guarantees for arbitrary signal distributions. Using this, we then derive
a novel strategy, \textbf{Christoffel-DPS}, for OSP in the setting of
DPS. Our specific contributions are:
\begin{enumerate}[leftmargin=1.5em,label=(\roman*)]
 \item A novel theoretical framework for recovery with \textit{arbitrary}
 (in particular, generative) priors that does not require Gaussian
 assumptions and applies to \textit{any} linear measurements.
 \item A novel OSP strategy, \textbf{Christoffel sampling},
 that is theoretically optimal for \textit{any} given prior.
 \item A practical implementation of Christoffel sampling in DPS,
 termed \textbf{Christoffel-DPS}, with both offline and online
 variants, for full state reconstruction from point measurements.
 \item Experiments on a series of scientific datasets showing the
 benefit of Christoffel-DPS over other OSP strategies for generative
 priors. A typical experimental result is shown in Figure \ref{fig:F1}.
\end{enumerate}

Our OSP strategy is fundamentally different to classical OED
criteria. Where A-, D- and E-optimal designs minimize the average
size of the posterior covariance ellipsoid -- in other words, they consider  how strongly a
measurement set collapses the posterior towards a mean state
-- Christoffel sampling controls the worst-case ratio between
measurement energy and signal energy across the secant set of the
support of the prior. It therefore asks how reliably a measurement
set \emph{identifies} between arbitrary candidate signals. The latter
is well-defined for arbitrary distributions and does not require
linearity or Gaussian assumptions. The shift in objective, from
posterior collapse to identifiability under noise, is what allows our
framework to remain meaningful when the posterior is non-Gaussian,
multi-modal or supported on a learned manifold. Christoffel functions
have recently emerged as powerful tools in deterministic
(non-Bayesian) recovery, using both linear \cite{cohen2017optimal} and nonlinear estimators \cite{adcock2023cs4ml}. In the former case, they are closely related to leverage scores \cite{chen2016statistical,ma2015statistical}. However, their use for Bayesian posterior sampling appears to be new.

\section{Setup and background on Christoffel sampling}
\label{s:cs-primer}

State reconstruction involves recovering an unknown signal $f^* : \Theta \rightarrow \bbR$ defined on a domain $\Theta$ (typically a subset of $\bbR^d$) from a sparse set of noisy sensor measurements
\bes{
y_i = f^*(\theta_i) + n_i,\quad i = 1,\ldots,m,
}
where $n_i \sim_{\mathrm{i.i.d.}} \cN(0,\sigma^2)$. OSP is the problem of choosing sensor locations $\theta_1,\ldots,\theta_m \in \Theta$ so as to maximize the accuracy of the recovered state, which we denote as $\hat{f}$.
In this work, we focus on \textit{posterior sampling} techniques, especially DPS. Let $\cP$ be a prior signal distribution, typically, a generative prior. Then posterior sampling involves drawing $\hat{f} \sim \cP(\cdot | y , \theta)$, where $y = (y_i)^{m}_{i=1}$ and $\theta = (\theta_i)^{m}_{i=1}$. Hence the OSP goal in posterior sampling is to maximize the fidelity of $\hat{f}$ to $f^*$ via the choice of $\theta$. 
To clarify, the recovery objective is that of generating $\hat{f}$ to minimize $L^2$ error between $f^*$ and $\hat{f}$, we are not directly optimizing for a better posterior $\cP( \cdot | y, \theta)$ in some probability metric. However, we can relate the two objectives by observing that if $f^* \sim \cP(\cdot | y , \theta)$, then $f^*$ and $\hat{f}$ are i.i.d conditional on the data and share expectation $\mu$, so the cross terms vanish and $$\mathbb{E}\left\|\hat{f}-f^*\right\|^2=2 \operatorname{tr} \operatorname{Cov}(f \mid y, \theta).$$ Then by choosing $\theta$ to minimize the error we also shrink the variance of the posterior and focus it on the mode containing $f^*$ -- see Fig.\ref{fig:F1}.

Now consider the domain $\Theta$ equipped with some measure $\rho$.
The key object in our work is the \textit{Christoffel function} $K(\cP)$ of $\cP$, defined as
\bes{
K(\cP) : \theta \in \Theta \mapsto \sup_{f \in \bbS(\cP)} | f(\theta) |^2 \in \bbR.
}
Here $\bbS(\cP)$ is the \textit{secant set} of $\mathrm{supp}(\cP)$
\be{
\label{secant-set}
\bbS(\cP) = \left \{ \frac{f_1 - f_2}{\nm{f_1-f_2}} : f_1 \neq f_2 \in \mathrm{supp}(\cP)\right \}
}
and $\nms{\cdot}$ is the $L^2_{\rho}(\Theta)$-norm. The Christoffel function is closely related to the concept of \emph{identifiability}: two distinct signals $f_1, f_2 \in \mathrm{supp}(\cP)$ should be distinguishable from their sensor values, as otherwise there is little hope of recovery. A sensor location $\theta$ is informative for this pair if the local difference $|f_1(\theta)-f_2(\theta)|^2$ is large compared to the global difference $\|f_1-f_2\|^2$ meaning their difference is then concentrated at $\theta$. At the other extreme, if $f_1(\theta)=f_2(\theta)$, this sensor location gives no information about the pair. The Christoffel function captures how informative a sensor location is towards identifiability of signals from $\cP$ and addresses the worst case scenario. Christoffel sampling, developed later, uses $K(\cP)$ to guide placement. By contrast, Fig.~\ref{fig:F1}, row 1, shows a pair that random sensors do not distinguish: the reconstruction differs from $f^*$ but has similar sensor values.

\section{Theoretical setup and results}\label{s:theory}

We now describe a general framework for theoretically-OSP. We base this on \cite{adcock_how_2025,jalal2021instance}, which introduced a framework for Bayesian recovery, but restricted to linear measurements between finite-dimensional vector spaces, and in the latter case, Gaussian measurements (which are not applicable to state reconstruction). A key generalization we develop here is the extension from problems formulated in $\bbR^N$ to arbitrary Hilbert spaces. Further, for technical reasons we elaborate in \S \ref{s:proofs}, the framework in these prior works is not suitable to OSP, since it assumes independence between the measurements and the noise. We relax this assumption, leading to the novel theoretically-OSP Christoffel sampling strategy discussed later. Finally, we note that our setup is broader than the state reconstruction problem defined in \S \ref{s:cs-primer}, in that it allows for sensing with arbitrary linear functionals, rather than just pointwise evaluations of the state.

\subsection{Setup}

Let $\bbX$ be a Hilbert space and $\bbX_0 \subseteq \bbX$ be a normed
vector subspace, termed the \textit{object space}. We consider arbitrary Hilbert spaces in this context so as to keep our framework discretization agnostic. Let $\cR,\cP$ be two probability distributions on $\bbX$ that take values
in $\bbX_0$ almost surely, termed the \textit{real} and
\textit{approximate} distributions, respectively. Our goal is to
recover an unknown $f^* \sim \cR$ from its measurements by sampling
from the posterior based on $\cP$. Note that, typically, $\cP$ is a generative prior that has been trained on a dataset drawn from $\cR$. However, this is not a requirement of our theory.

We now define our sensing model.
Let $(\Theta,\cT,\rho)$ be a measure space. We term $\Theta$ the \textit{sensor parametrization set}: in brief, $\theta_0 \in \Theta$ parametrizes a sensor, telling it what measurement to take. Next, we define a \textit{sensing operator} as an injective map
\bes{
L : \Theta \rightarrow \cB(\bbX_0),
}
where $\cB(\bbX_0)$ is the set of bounded, linear functionals $\bbX_0 \rightarrow \bbR$. Now let $\mu$ be such that $(\Theta,\cT,\mu)$ forms a probability space. We term $\mu$ the \textit{sampling measure}. With this in hand, let $f^* \in \bbX_0$ be the object to recover. To measure $f^*$, we draw $\theta_1,\ldots,\theta_m \sim_{\mathrm{i.i.d.}} \mu$ and consider noisy measurements
\bes{
y = (y_i)^{m}_{i=1} \in \bbR^m,\quad \text{where }y_i = L(\theta_i)(f^*) + n_i,\ i = 1,\ldots,m,
}
and $n_i \sim \cN(0,\sigma^2)$ is measurement noise that is independent of the $\theta_i$. For convenience, we write
\be{\label{data-model}
\theta = (\theta_i)^{m}_{i=1},\quad n = (n_i)^{m}_{i=1},\quad L(\theta)(f^*) = (L(\theta_i)(f^*))^{m}_{i=1},\quad  y = L(\theta)(f^*) + n.
}
Finally, we assume that $L$ is \textit{nondegenerate with respect to $\cP$}. Namely, there exist constants $0 < \alpha \leq \beta < \infty$ such that
\be{
\label{nondegeneracy}
\alpha \nm{f}^2 \leq \int_{\Theta} | L(\theta)(f) |^2 \D \rho(\theta) \leq \beta \nm{f}^2,\quad \forall f \in \bbS(\cP),
}
where $\bbS(\cP)$ is the secant set \ef{secant-set}. Here and elsewhere $\nm{\cdot}$ is the norm on $\bbX$.
Note that this assumption is mild, and essentially states that the energy of any signal $f$ should be approximately preserved, on average, by the sensing operator.

\rem{
[Typical sensing operators]
In this paper, we primarily consider state reconstruction from point samples. This problem can be formulated in the above framework by letting $\bbX = L^2_{\rho}(\Theta)$ be the space of $L^2$ functions on some domain $\Theta$,  $\bbX_0 = C(\overline{\Theta})$ and $L : \Theta \to \cB(\bbX_0)$ with $L(\theta)(f) = f(\theta)$, $\forall f \in \bbX_0$. 
Notice that $L$ satisfies \ef{nondegeneracy} with $\alpha = \beta = 1$. 
However, while not the focus of this work, we note in passing that our framework can easily handle other types of recovery problems. This is important to consider as many relevant problems utilize different sensing operators. One straightforward modification is the gradient-augmented problem, where $L(\theta)(f) = (f(\theta) , \nabla f(\theta) )^{\top}$. This type of sampling arises in various scientific and engineering applications. Another application is to inverse PDE problems, where the function $f$ is some state corresponding to a PDE (e.g., an initial condition or inhomogeneity) and $L(\theta)(f) = u(\theta)$ evaluates the PDE solution $u$ corresponding to $f$ at $\theta$. Thus, our framework can be readily applied to (linear) inverse PDE problems.
}

\subsection{Main theoretical result}

With this in hand, we now consider how to choose $\mu$ optimally so as to minimize the number of measurements/sensors $m$ required to recover an unknown $f^* \sim \cR$ accurately by sampling from the approximate posterior $\hat{f} \sim \cP(\cdot | y , \theta )$, where $\theta = (\theta_i)^{m}_{i=1}$. Specifically, given $t > 0$, we shall estimate $p = \bbP [ \nm{f^* - \hat{f} } \geq t  ]$, where the probability is taken with respect to all variables, i.e., $f^* \sim \cR$, $\theta \sim \mu^{\otimes m}$, $e \sim \cN(0,\sigma^2 I_m)$ and $\hat{f} \sim \cP(\cdot | y , \theta)$, where $y = L(\theta)(f^*) + e$. Note, we do not directly optimize the posterior, but rather parameters $\theta_i$ so that the $L^2$ error of $f^*$ and $\hat{f}$ is minimized with high probability. To state our main result, we first require the following.

\defn{[Approximate covering number]
\label{d:approx-cov-num}
Let $(X, \cF, \cP)$ be a probability space and $\delta, \eta \geq 0$. The \textit{$\eta, \delta$-approximate covering number of $\cP$} is defined as 
\bes{
    \mathrm{Cov}_{\eta, \delta}(\cP) = \min \left\{ k \in \bbN : \exists \{ x_i \}^{k}_{i=1} \subseteq \mathrm{supp}(\cP),\ 
    \cP \left(\bigcup_{i=1}^k B_{\eta}(x_i) \right) \geq 1 - \delta \right\}.
}
}
Approximate covering numbers quantify the minimum number of balls of radius $\eta$, to cover $1 - \delta$ of the mass of $\cP$. At a high level, this characterizes the complexity of the prior $\cP$ and therefore plays a key role in the number of measurements needed for recovery. For further discussion and examples of covering numbers for typical distributions, see \cite{adcock_how_2025, jalal2021instance}. We consider arbitrary priors $\cP$ to cover all possibilities of priors used in posterior sampling including Gaussian priors \cite{huan2024optimal}, hand-crafted priors (such as sparse vectors) \cite{adcock_how_2025, dashti2017bayesian}, and generative priors \cite{chung2023diffusion}.

\thm{
\label{t:sensor-distribution}
Let $1 \leq p \leq \infty$, $0 < \delta \leq 1/4$, $\varepsilon,\eta > 0$ and consider the above setup. Suppose that $\mu \ll \rho$, $ \D \mu / \D \rho > 0$ $\mu$-a.s., let $w(\theta) = \left ( \D \mu(\theta) / \D \rho \right )^{-1}$ and define
\bes{
w_{\min} = \mathrm{essinf}_{\theta \sim \rho} w(\theta),\quad w_{\max} = \mathrm{esssup}_{\theta \sim \rho} w(\theta).
}
Suppose that
\be{
\label{Wp-sigma-cond}
W_p(\cR , \cP) \leq \varepsilon\quad \text{and}\quad \sigma \geq \frac{m \varepsilon}{w_{\min} \delta^{1/p+1/2}}.
}
Then
\bes{
\bbP [ \nm{f^* - \hat{f} } \geq (32 \max \{ w_{\max} , 1/\alpha , \beta \} + 2)(\eta + \sigma)  ] \lesssim \delta ,
}
provided
\be{
\label{num-meas-general-m}
m \geq c(\alpha,\beta) \cdot \left ( \kappa_{w}(\cP , L ) + 1 \right ) \cdot \left ( \log \mathrm{Cov}_{\eta,\delta}(\cP) + \sqrt{w_{\max}} + \log(1/\delta) \right ),
}
where $c(\alpha,\beta) > 0$ depends on $\alpha,\beta$ only and
\be{
\label{kappa_w_def}
\kappa_{w}(\cP , L) = \esssup_{\theta \sim \rho} \Big \{ w(\theta) \sup_{f \in \bbS(\cP)} | L(\theta)(f)|^2  \Big \}.
}
}
This theorem provides a \textit{nonasymptotic} guarantee for successful Bayesian recovery with high probability with \textit{arbitrary} priors $\cP$. Further, it addresses the realistic setting where the true signal $f^*$ is drawn from some unknown real distribution which is close to the (typically learned) distribution $\cP$ that is used as the prior. In particular, Theorem \ref{t:sensor-distribution} states the error $\nm{f^* - \hat{f}}$ is within a constant of $\eta + \sigma$, where $\eta$ is an accuracy parameter and $\sigma$ is the noise level, with probability at least $1-\delta$, provided the number of sensors satisfies \ef{num-meas-general-m}. This latter estimate states that the sample complexity is dictated by $\log \mathrm{Cov}_{\eta,\delta}(\cP)$ multiplied by $\kappa_w(\cP,L)$. The former is a measure of the complexity of $\cP$: more complex priors require more measurements for successful recovery. The latter dictates how the prior interacts with the sensing operator $L$ and the sampling distribution $\mu$ (see the term $w(\theta)$).

Theorem \ref{t:sensor-distribution} holds for almost arbitrary sampling measures $\mu$. In order to identify a theoretically-optimal choice $\mu^{\star}$, we choose $w$ that minimizes the right-hand side of \ef{num-meas-general-m}. Assuming measurability and recalling that $w$ must satisfy $\int w(\theta)^{-1} \D \rho(\theta) = 1$ since $\mu$ is a probability measure, the optimal choice is given by
\bes{
w(\theta)^{-1} = \frac{K(\cP,L)(\theta)}{C(\cP,L)}
}
where $K(\cP,L)$ is the \textit{(generalized) Christoffel function} \cite{adcock2023cs4ml}
\bes{
K(\cP,L) = \sup_{\substack{f \in \bbS(\cP)}} |L(\theta)(f)|^2 ,\qquad C(\cP,L) = \int_{\Theta} K(\cP,L)(\theta) \D \rho(\theta) .
}
This choice of $w$ minimizes \ef{kappa_w_def}, yielding the minimal value $\kappa_w(\cP,L) = C(\cP,L)$.
However, it is also convenient to choose $w(\theta)$ that is not too large, which will allow one to remove the dependence on $w_{\max}$ in the above bounds. Fortunately, this can be easily done, by choosing
\bes{
w(\theta)^{-1} = \frac{K(\cP , L)(\theta)} {2C (P,L)} + \frac{1}{2}.
}
In this case, one has $w_{\max} \leq 2$, while $\kappa_w(\cP,L) \leq 2 C(\cP,L)$ is within a factor of $2$ of being optimal. Using this, we now choose $\mu^{\star}$ as follows:
\be{
\label{optimal-mu}
\D \mu^{\star}(\theta) = \left ( \frac{K(\cP , L)(\theta)} {2C (P,L)} + \frac{1}{2} \right ) \D \rho(\theta).
}
We summarize this discussion in the following result.

\cor{
[Theoretically-optimal sampling]
\label{cor:christoffel-sampling}
Assume the setup of the previous theorem, with $\mu = \mu^{\star}$ given by \ef{optimal-mu}. Then
\bes{
\bbP [ \nm{f^* - \hat{f} } \geq (32 \max \{  2,1/\alpha , \beta \} + 2)(\eta + \sigma)  ] \lesssim \delta ,
}
provided
\be{
\label{m-cond-CPL}
m \geq c(\alpha,\beta) \cdot \left ( C(\cP , L) + 1 \right ) \cdot  \left ( \log \mathrm{Cov}_{\eta,\delta}(\cP)  + \log(1/\delta) \right ).
}
}

This result establishes a theoretically-optimal (random) sensor placement strategy at a very high level of generality, in that it can be applied to arbitrary distributions $\cP$ and sampling operators $L$. We term the choice $\mu = \mu^{\star}$ as \textit{Christoffel sampling}. See App. \ref{app:christoffel-dicussion} for further discussion.

\section{Christoffel-DPS}
\label{s:cdps}

We now describe our algorithms for practical implementation of Christoffel sampling in the setting of DPS, which we term Christoffel-DPS, with both online and offline variants. Throughout, we work in the discrete setting and we consider $L$ as the pointwise evaluation operator, i.e., $L(j)(x) = x_j$ for $x \in \bbR^N$. Notably, computing $K(\cP,L)$ for a generative prior involves a supremum over $\bbS(\cP)$, which is typically intractable. We propose two practical instantiations to handle this.

\begin{itemize}
\item \textbf{Offline} (\S\ref{ss:offline}): estimate $K$ from a finite
secant set drawn once, before sampling.
\item \textbf{Online} (\S\ref{ss:online}): recompute $K$ on a live
DPS ensemble at scheduled \textit{drift events}, allowing for adaptive sensor placements based on measurements obtained during prior drift events.
\end{itemize}

In practice, the state reconstruction problem is often formulated over a finite grid. Let $\{ \xi_i \}^{N}_{i=1} \subset \Theta$ be a fixed grid of nodes and identify the vector $x^* \in \bbR^N$ with $f^*$ via $x^*_i = f^*(\xi_i)$. In this case, $K(\cP)$ becomes a score function over $\{1,\ldots,N\}$ with $K(\cP)(j) = \sup_{f \in \bbS(\cP)} | f(\xi_j) |^2$. Here, we also adopt the notation $S \in \bbR^{m \times N}$ for a row-selection matrix comprising $m$ rows of the identity $I_N$, which corresponds to the selected sensors, and express the measurements as
\be{
\label{finite-state-model}
y_S = S x^* + n,\quad n \sim \cN(0,\sigma^2 I_m).
}

\subsection{Diffusion posterior sampling (DPS)}\label{ss:dps}

We first briefly review DPS \citep{chung2023diffusion} and introduce notation we will use later.
For convenience we present the construction in finite dimensions and with sensor measurements given by \ef{finite-state-model}; the extension to
function-space diffusion models follows
\citep{pidstrigach2024infinite} via a trace-class
covariance operator $C$ replacing the identity. Let $\cP$ be a prior induced by a diffusion model on $\bbX = \bbR^N$ and let
$D_\theta(x,\sigma)\approx\bbE[x_0\mid x_t = x]$ denote the corresponding denoiser
trained by score matching on samples, with associated
score $s(t,x) = -\sigma^{-2}(t)\bigl(x - D_\theta(x,\sigma(t))\bigr)$.
The variance-exploding
forward SDE \citep{karras_elucidating_2022} is
\bes{
dx_t = g_t \sqrt{C}\, dW_t,\qquad x_0\sim\cP,\qquad g_t = \sqrt{d\sigma^2/dt},
}
with transition law $\cN(x_0,\sigma^2(t)C)$, and the reverse SDE
$dz_t = g_{T-t}^2\,s(T-t,z_t)\,dt + g_{T-t}\sqrt{C}\,dW_t$ starting from
$z_0\sim\cN(0,\sigma^2(T)C)$ yields $z_T\sim\cP$. 
DPS
samples from the posterior given $y_S = Sx^* + \eta$ by augmenting
the unconditional score with a Tweedie-approximated measurement
gradient,
\bes{
\nabla\log h^{y}(t,x) \;\approx\; -\nabla_x\,\Phi\!\bigl(D_\theta(x,\sigma(t)),\,y_S\bigr),
\qquad
\Phi(x_0,y) = \tfrac{1}{2\sigma_\eta^2}\nm{Sx_0 - y}^2.
}
In the $\sigma(t)\!\to\!0$ limit this reduces to the Kalman update
with gain set by the local denoiser Jacobian -- in particular, no assumption on $\cP$
beyond access to $D_\theta$ is required. 

\subsection{Offline Christoffel-DPS}\label{ss:offline}

We first describe two offline Christoffel-DPS variants.
In both cases, we commence with a set of snapshots $\{x^{(n)}\}_{n=1}^{M}\subset\mathrm{supp}(\cP)$, either training data or unconditional samples drawn once from $\cP$.

\textbf{Christoffel-DPS.}\ This approach is a near-direct application of the theory in \S \ref{s:theory}. We define the \emph{empirical secant set}, the finite-$M$ analogue of the
secant set $\bbS(\cP)$ from \S\ref{s:theory}, as
\bes{
\bbS_M =  \left \{ \frac{x^{(n)} - x^{(n')}}{\nm{x^{(n)} - x^{(n')}}} : n\neq n' \right \} \subset  \bbR^N
}
and then compute the empirical Christoffel function $\widehat{K}_M$ by replacing the supremum over $\bbS(\cP)$ with the maximum over $\bbS_M$, giving
\be{\label{K-emp}
\widehat{K}_{M}(j) = \max_{x \in \bbS_M} |x_j|^2 ,\quad j\in\{1,\ldots,N\}.
}
Note that  $\widehat{K}_M \to K$ as $M \rightarrow \infty$. With \ef{K-emp} in hand, Christoffel-DPS proceeds by selecting $m$ sensors i.i.d.\ according to the discrete probability distribution on $\{1,\ldots,N\}$ with weights \ef{K-emp}.

\rem{[OSP with Christoffel-DPS is not MAP seeking]
The offline Christoffel function depends only on our prior, $\mathcal{P}$ and the sensor operator $L$. It is computed before any measurement exists, so it cannot depend on $y$ or $f^*$, hence it cannot be selecting placements that make $f^*$ the MAP, since it never sees $f^*$.
}

\textbf{Greedy Christoffel-DPS.}\ The theoretical analysis in \S \ref{s:theory} demonstrates the fundamental nature of the secant set $\bbS_M$. In this variant, we implement an alternative to i.i.d.\ sampling that builds an OSP in a greedy manner. This is reminiscent of classical OED design criteria, but differs starkly, as it works directly on the empirical secant set $\bbS_M$ corresponding to a highly nonlinear prior $\cP$.

Let $X \in \bbR^{N \times M}$ be the matrix of mean-adjusted snapshots $x^{(n)} - x^{(\mathrm{avg})}$, where $x^{(\mathrm{avg})} = \frac1M \sum^{M}_{n=1} x^{(n)}$. Let $S \in \bbR^{m \times N}$ be row-selection matrix and notice that
\be{
\label{information-matrix}
 S X  (S X)^{\top}= \frac{1}{M^2} \sum_{n,n',n''} \left ( S \left ( x^{(n)} - x^{(n')} \right ) \right ) \left ( S \left ( x^{(n)} - x^{(n'')} \right ) \right )^{\top} 
}
Using ideas from \cite{karnik_pysensors_2025,manohar_data-driven_2018}, we now construct $S$ in a greedy manner to maximize the information gain in this matrix. In practice, this is done by column-pivoted QR factorization; see App.\ \ref{app:offline}.

Note that approach does not directly work with the empirical secant set $\bbS_M$, due to computational resources ($|\bbS_M| = M (M-1)$ in general). Instead, it uses the empirical average $x^{\mathrm{avg}}$ to effect a cheap approximation to pairwise differences, as seen in \ef{information-matrix}, retaining the original dataset size $M$. Normalization is also ignored, as the averaged snapshots $x^{(n)} - x^{(\mathrm{avg})}$ have similar norms in practice.

\rem{[Relation to POD-based OED]
A standard approach to Bayesian-OED for state estimation involves first computing a POD (proper orthogonal decomposition). Here one computes an orthonormal basis $\{ v^{(1)},\ldots,v^{(r)}\} \subset \bbR^N$ from the set of snapshots $\{ x^{(n)} \}^{M}_{n=1}$, then forms the design matrix $V \in \bbR^{N \times r}$ from this basis and applies an OED criterion (e.g., A- or D-optimality, or some heuristic) on $V$, potentially with a data-driven regularization, as in \cite{manohar_data-driven_2018,chaturantabut_nonlinear_2010,drmac_new_2016,karnik_pysensors_2025,klishin_data-induced_2025}. However, this differs significantly from greedy Christoffel-DPS, as the POD basis is, in effect, approximating the support of $\cP$ by a linear $r$-dimensional subspace. Thus it may fail to capture the geometry of $\cP$ when using complex, highly non-Gaussian priors. By contrast, greedy empirical Christoffel works on the secant set, which better captures the geometry of $\cP$.
}

\subsection{Online ensemble Christoffel-DPS}\label{ss:online}

A benefit of generative recovery is that the sensor placement need not be fixed in advance in an offline fashion, but rather updated online by using the fact that the Christoffel function can be repeatedly re-estimated during the diffusion process conditioned on the current measurements.
To achieve this, we run an ensemble of $N_e$ DPS chains
$\{z^{(i)}_t\}_{i=1}^{N_e}$ sharing the current row-selector matrix $S$ and
measurement vector $y_S$, and schedule $D$ \emph{drift events} at
noise levels $\sigma(t_1)>\cdots>\sigma(t_D)$ along the reverse
diffusion.
At drift event $t_d$ each chain emits its Tweedie point estimate
$\hat x^{(i)}_0 = D_\theta(z^{(i)}_{t_d},\sigma(t_d))$, and the
empirical Christoffel function is evaluated on the secant set of the resulting ensemble:
\be{\label{K-online}
\widehat{K}_{t_d}(j) = \max_{i\neq i'} 
\frac{\bigl | \hat x^{(i)}_{0,j} - \hat x^{(i')}_{0,j}\bigr |^2}
{\nm{\hat x^{(i)}_0 - \hat x^{(i')}_0}^2},
\qquad j = 1,\ldots,N.
}
We then use this to relocate the sensors by sampling i.i.d.\ with probabilities proportional to $\widehat{K}_{t_d}(j)$.
The new entries of $y_S$ are
then read, chains whose measurement residual exceeds a quantile
threshold are pruned, and the survivors resume under the new conditional
score until the next drift event.  After the $D$th event the
ensemble is collapsed either by selecting the chain with the smallest
measurement residual or by averaging over survivors. We detail this algorithm in App.\ \ref{app:online}.

\section{Experimental Results}
\label{s:experiments}

We evaluate Christoffel-DPS on three DPS benchmarks: the fluidic pinball (GRIFDIR \cite{rowbottom_grifdir_2026}), the Darcy forward problem (DiffusionPDE \citep{huang_diffusionpde_2024}), and Kolmogorov flow \citep{amoros-trepat_guiding_2026}. For each, we take a trained denoiser, and run 10 random seeds of DPS guidance with the reverse-diffusion sampler and generate the row-selector $S$ once offline using the following placement strategies: random; A-, D- and E-optimal POD \citep{joshi_sensor_2009}, SSPOR using PySensors 2.0 \cite{karnik_pysensors_2025}; the offline Christoffel-DPS variants of \S\ref{ss:offline} and the online ensemble Christoffel-DPS method of \S \ref{ss:online}. Each (strategy, number of sensors $m$) cell is averaged over the random seeds and reported as mean $\pm$ std of the relative $L^2$ error $\nm{\hat x - x^*}_2 / \nm{x^*}_2$. In each experiment, our primary metric is to minimize the $L^2$ error between $x^*$ and $\hat{x}$. Code is available at \url{https://github.com/JRowbottomGit/Christoffel-DPS}.
\begin{wrapfigure}[13]{r}{0.5\linewidth}
  \centering
  \vspace{-10pt} 
  \includegraphics[width=\linewidth]{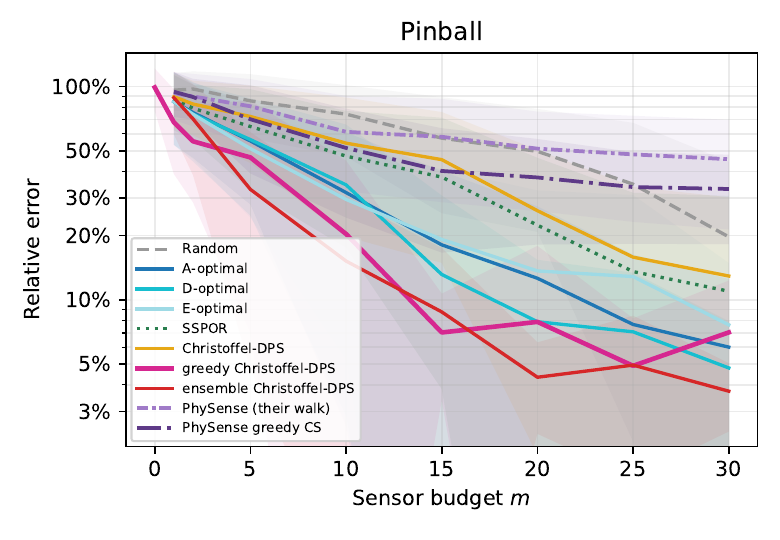}
  \caption{\textbf{Pinball: $m$-convergence.} Relative $L^2$ error vs sensor budget $m$; bands one std. Purple: PhySense comparison (Sec.~\ref{sec:physense}).}
  \label{fig:F4_pinball}
\end{wrapfigure}

\subsection{Pinball problem with GRIFDIR}
\label{ss:pinball}

Our first experiment is on the Pinball problem \cite{tomasetto2025reduced}. We instantiate Christoffel-DPS in the infinite-dimensional setting
via GRIFDIR \citep{rowbottom_grifdir_2026}, a function-space EDM
diffusion model on an unstructured domain with continuous
FEM-continuous graph kernel layers and a multiscale graph and latent space transformer backbone.  The forward noise is a Gaussian random field with covariance operator $C$, and the same $C$ broadcasts the DPS measurement-gradient at
inference: with likelihood potential
$\Phi(x_0, y_S) = \tfrac{1}{2\sigma_\eta^2}\nm{S x_0 - y_S}^2$, the
conditional-score correction (cf.\ \S\ref{ss:dps}) is approximated by $\nabla\log h^{y}(t,x) \approx  -\nabla_x \Phi \bigl(D_\theta(x,\sigma(t)),y_S\bigr)$.

The denoiser is trained on snapshots of the scalar field $c$ governed by the advection--diffusion equation $\partial_t c + v(\mu;x)\!\cdot\!\nabla c - D\Delta c = 0$, where $v(\mu;x)$ is the steady RANS solution around three rotating pinball cylinders with three rotation rates $\mu$ selected uniformly from $[-5,5]$.  Snapshots live on the unstructured FEM mesh of the simulation, so $S$ is implemented to select mesh nodes.  Figure~\ref{fig:F4_pinball} shows that the
online ensemble Christoffel-DPS reaches the error of the random and classical baselines (A-, D-, E-optimal POD, SSPOR) at roughly half the sensor budget, and the offline greedy Christoffel-DPS variant matches or exceeds the strongest classical
baseline at almost all budgets. Full experimental details are given in App.~\ref{app:experiments-pinball}. In the appendix, Fig.~\ref{fig:scorefield} plots the Christoffel function over the Pinball dataset and shows how greedy Christoffel-DPS first selects symmetric points that subdivide the eight octants of the cylinders' angular rotation, maximising identifiability so the posterior collapses within a handful of sensors.

\subsection{Darcy flow with DiffusionPDE}
\label{ss:darcy}

\begin{wrapfigure}{r}{0.5\linewidth}
  \centering
  \vspace{-15pt} 
  \includegraphics[width=\linewidth]{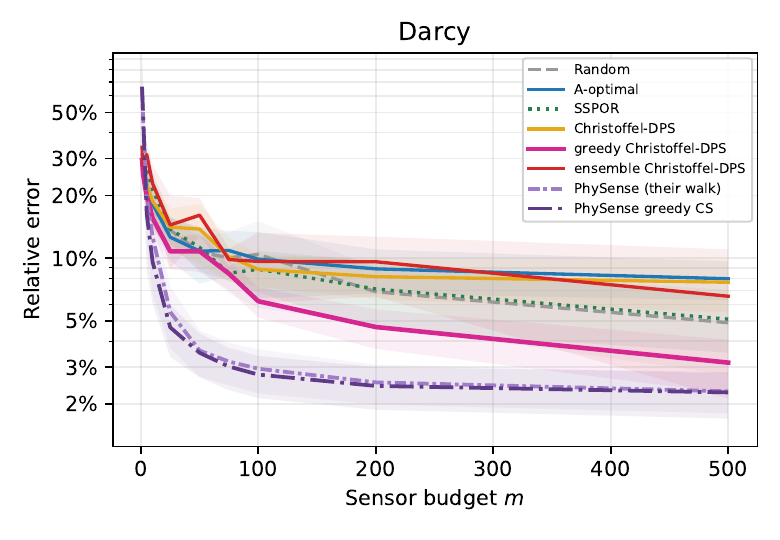}
  \caption{\textbf{Darcy flow: $m$-convergence.} Relative $L^2$ error vs sensor budget $m$; bands one std. Purple: PhySense comparison (Sec.~\ref{sec:physense}).}
  \vspace{-15pt} 
  \label{fig:F7_darcy}
\end{wrapfigure}

Our second experiment is full state reconstruction on Darcy flow \cite{huang_diffusionpde_2024, li2021fourier}. We use the joint $(a, u)$ DhariwalUNet \citep{karras_elucidating_2022} trained on the steady Darcy equation $-\nabla \cdot (a\nabla u) = f$ on a uniform $128 \times 128$ grid with binary conductivity $a \in \{3, 12\}$ and pressure $u$.  Recovery uses the Karras-EDM DPS sampler: at every reverse Heun step the measurement gradient $\nabla_{x_t}\nm{y_S - S\hat x_0}^2$ is taken via auto-grad through the denoiser and added to the unconditional score.  

With $128\times128=16,384$ pixels and 500 Heun steps as our default hyper-parameters this data was comparatively expensive compared to pinball which has roughly 7,500 mesh nodes and used 50 Heun denoising steps. For practical reasons we reduced the ensemble size from $N_e=20$ to 10 and reverse-diffusion budget from $2000$ to $500$ Heun steps. This may explain the slight under performance of ensemble Christoffel in this experiment compared to Pinball and Kolmogorov flow, which we discuss in the following section.

Finally, OSP methods that require a POD basis are implemented with independent $a,u$ placement strategies. The D-/E-optimal methods collapse due to rank-deficiency of the low-frequency field $u$ and are omitted from Fig.~\ref{fig:F7_darcy}. Full experimental details are given in App.~\ref{app:experiments-darcy} including Table~\ref{tab:physense_darcy} that reports them alongside Tikhonov-regularized variants (D,E-OPT-REG) with prior $\Sigma_0 = \mathrm{diag}(\lambda) + \epsilon I$, $\epsilon = 10^{-4}$.

\subsection{Kolmogorov flow with physics-constrained masked diffusion}

For our third benchmark we use the U-Net DDPM-style denoiser of
\citet{amoros-trepat_guiding_2026}, trained on vorticity snapshots
of the 2D Kolmogorov flow. 
The model uses a novel ConFIG gradient-projection step to form consensus on score and guidance gradients and rather than backpropagating a measurement gradient at every reverse step, recovery uses a latent-mask-blending DPS sampler, the denoiser's prediction at observed indices is overwritten with the RBF-broadcast measurements.  The reverse loop runs entirely in \texttt{torch.no\_grad}, so the per-step cost is dominated by a single forward denoiser pass per chain and the ensemble overhead is modest; the online ensemble Christoffel-DPS therefore runs at the full $N_e = 20$ chains and the unmodified reverse-diffusion budget (in contrast to the slimmed Darcy configuration of \S\ref{ss:darcy}). 
Experimental results and full experimental details are given in App.~\ref{app:experiments-kmflow}.

\subsection{Comparison against optimized sensor placement of PhySense}\label{sec:physense}

PhySense~\cite{ma_physense_2025} couples sensor placement with a transformer based conditional flow matching model. The sensor locations are network inputs and the model is trained over randomly selected locations at sensor-counts in the range ($m \in [25,400]$). A secondary stage then walks the sensors from initial SSPOR or random placements to their optimal locations, by gradient descent of the reconstruction error with respect to the sensor coordinates through the frozen model. The authors show the flow matching loss controls the A-optimality criteria and as such we expect the sensors to converge on the A-optimal floor. 

We retrain PhySense on each of our three benchmarks. When the flow matching model is provided with sensor locations generated by greedy-Christoffel once offline, that were neither seen in training or related to A-optimal objective, they still outperform the optimised walked placements on Darcy and Pinball, while on Kolmogorov flow the two trade places within a few percent. Curves for both are included in Fig~\ref{fig:F4_pinball}, Fig~\ref{fig:F7_darcy};
training details and full tables, including a random-initialization walk, are in Appendix~\ref{app:experiments}. We note the conditional model is restricted to budgets near its training range, whereas the DPS prior is trained once and queried at any $m$.

\section{Discussion and limitations}
\label{s:discussion-limitations}

We presented a framework for theoretically-OSP based on Christoffel functions in the setting of posterior sampling. Using this theory, we introduced Christoffel-DPS, a novel algorithm for OSP for sparse sensor reconstruction tasks using DPS.
We achieved this by empirically estimating the Christoffel function, in both online and offline fashions, and exploiting it in both random and greedy sensor placement strategies.
Our methods differ from classical OED methods, as our sensor placement strategies are specifically designed for non-linear, non-Gaussian priors such as those arising in DPS. On datasets such as 
Pinball and Kolmogorov flow, Christoffel-DPS, particularly greedy Christoffel-DPS, reach classical baselines at approximately half the sensor budget compared to other methods. Even in the Darcy flow experiment, greedy Christoffel-DPS reaches a lower baseline and outperforms classical methods on low sensor budgets. Non-greedy Christoffel-DPS tends not to perform as well, which we believe is due to the expensive and difficult nature of estimating the empirical secant set.

We finish off by discussing limitations and avenues for future work. As a first point, while our theory allows for arbitrary types of linear measurements, we have only considered state reconstruction from point evaluations in our experiments. Future work on, for instance, gradient augmented-data \cite{adcock2023cs4ml} have yet to be explored. Outside of state estimation, inverse PDE problems are another important class of problems not addressed in this work. While our theory can be applied directly to linear PDEs, our algorithms have not yet been applied to such problems. For both theory and algorithms, however, nonlinear PDEs remain a more substantial challenge. We are currently working on extending our framework to nonlinear inverse problems. Next, it is notable that while Christoffel-DPS outperforms classical OED methods in many cases, their benefits can be mild on datasets/problems which are approximately Gaussian and linear. We intend to explore other highly non-linear, non-Gaussian datasets in future works. On the algorithmic side, our methods require approximation of the secant set, which can be expensive. We mitigated this issue with greedy Christoffel-DPS, but it remains to be seen if there are more efficient ways to estimate this set. Finally, our online algorithm is expensive, requiring ensembles with DPS. In particular the guidance term of DPS requires automatic differentiation at every reverse step, which is expensive at inference. We remark that this problem is intrinsic to expensive DPS strategies, in experiment 3 we found a solution to this using masked diffusion. Overall, the landscape of generative modelling is rapidly evolving, the state estimate techniques explored in this paper are well poised to benefit from any algorithmic or technological advances from the wider field. Our theory provides a principled and complementary partner for future OED research that is flexible to any new posterior sampling methods. 

\section*{Acknowledgements}
We thank Alexander Denker for helpful discussion and feedback. JR acknowledges support from the EU Horizon MSCA-SE under project REMODEL, ``Research Exchanges in the Mathematics of Deep Learning with Applications'' (grant agreement no. 101131557), and the UKRI AI Research Resource (AIRR) for compute on Dawn [ST/Z000890/1] and Isambard-AI [ST/AIRR/I-A-I/1023]. CBS acknowledges support from the Royal Society Wolfson Fellowship, the EPSRC advanced career fellowship EP/V029428/1, the EPSRC programme grant EP/V026259/1, the Wellcome Innovator Awards 215733/Z/19/Z and 221633/Z/20/Z, the EPSRC funded ProbAI hub EP/Y028783/1, and the European Union Horizon 2020 research and innovation programme under the Marie Sk{\l}odowska-Curie grant agreement REMODEL. BA acknowledges support from the Natural Sciences and Engineering Research Council of Canada (NSERC) through grants RGPIN-2026-04531, DGDND-2026-04531 and FRQ (Fonds de recherche du Qu\'ebec) -- Nature et Technologies through grant 359708. NH acknowledges support from an NSERC Canada Graduate Scholarship. The authors declare no competing interests.

\bibliographystyle{plainnat}
\bibliography{HMSAE,manual_refs}

\newpage 
\appendix

\begin{table}[p]
\centering
\begin{tabular}{|l|p{4.5in}|}
    \hline
    Notation & Meaning\\
    \hline
    $\cP, \cR$ & Arbitrary (Borel) distributions \\
    $\bbX$ & Hilbert space \\
    $\bbX_0$ & Object space \\
    $L$ & Sensing operator \\
    $\Theta$ & Parameter space (often sensor location space/physical domain) \\
    $\theta$ & Parameters in the parameter space (often sensor placement locations) \\
    $\rho$ & Base measure \\
    $\mu$ & Parameter measure \\
    $w(\theta)$ & Reciprocal of the Radon-Nikodym derivative of $\mu$ w.r.t $\rho$ at $\theta$\\
    $\bbS(\cP)$ & Secant set of distribution $\cP$ \\
    $K(\cP, L), K(\cP)$ & Christoffel function of distribution $\cP$ with respect to $L$\\
    $C(\cP, L)$ & Integral of $K(\cP, L)$ \\
    $\hat{K}$ & Empirical estimate of $K$ \\
    $\mathrm{Cov}_{\eta, \delta}(\cP)$ & $\eta, \delta$-approximate covering number of $\cP$ \\
    $n_i$ & Gaussian noise \\
    $f^*$ & Object of recovery \\
    $\hat{f}$ & Recovered object that approximates $f^*$ \\
    $\bbP$ & Probability of \\
    $\bbE$ & Expected value \\
    $\nm{\cdot}$ & $L^2$ norm (implicitly under measure $\rho$)\\
    $x^*, \hat{x}$ & Discretized variants of $f^*, \hat{f}$, respectively. \\
    \hline
\end{tabular}
\caption{Notation used throughout the paper.}
\label{tab:notation}
\end{table}

\section{Proofs}\label{s:proofs}

We now prove the main results of the paper. We do this by first presenting an abstract theoretical framework for accuracy and stability when solving Bayesian inverse problems in general Hilbert spaces. Theorem \ref{t:sensor-distribution} then follows as a corollary of this abstract framework. As mentioned, our approach is based primarily on \cite{adcock_how_2025,jalal2021instance}, but with substantial generalizations to, firstly, handle infinite-dimensional Hilbert spaces and, secondly, to provide a more flexible framework from which we can derive the theoretically-OSP strategy, i.e., Christoffel sampling. Specifically, these works considered only discrete problems in $\bbR^N$, which we now generalize to problems in arbitrary separable Hilbert spaces $\bbX$. Further, the abstract framework in \cite{adcock_how_2025,jalal2021instance} assumed independence between the measurement distributions and the noise distribution. For reasons we describe in Remark \ref{rem:dependence}, this assumption is violated in the OSP setting of Theorem \ref{t:sensor-distribution}. In this work, we generalize these prior works by allowing the sensing operator and noise to be dependent.

\subsection{Setup and objective}

Let $\bbX$ be a Hilbert space. Notice that this automatically implies that $\bbX$ is a Polish space, i.e., a separable completely metrizable Hilbert space. Let $\cR, \cP$ be Borel probability distributions on $\bbX$. Let $\cL(\bbX_0, \bbR^m)$ denote the space of continuous linear operators $\bbX \rightarrow \bbR^m$. Now let $\cM$ be a probability distribution on $\cL(\bbX , \bbR^m) \times \bbR^m$ with marginals $\cA$ and $\cE$. Now let $(A,e) \sim \cM$ and $f^* \sim \cR$ independently. Given measurements $y = A(f^*) + e$, our aim is to find conditions on $\cM, \cP, \cR$ such that, for small $t > 0$, 
\be{
\label{p-general-def}
p  : = \bbP_{f^* \sim \cR, (A,e) \sim \cM, \hat{f} \sim \cP(\cdot | y, A)} \{ \nm{f^* - \hat{f}} \geq t \}
}
is small. When understood, we use $\nm{\cdot}$ to denote either the Hilbert space norm on $\bbX$ or the Euclidean norm on $\bbR^m$.

\subsection{Key definitions}

We require several definitions. Let $\cA$ be as above and $D \subseteq \bbX$. Then a \textit{lower concentration bound for $\cA$} is any constant  $C_{\mathsf{low}}(t) = C_{\mathsf{low}}(t ; \cA, D) \geq 0$ such that 
\bes{
\bbP_{A \sim \cA} \{ \nm{A(f)} \leq t \nm{f} \} \leq C_{\mathsf{low}}(t ; \cA,D),\quad \forall f \in D.
}
Similarly, an \textit{upper concentration bound for $\cA$} is any constant $C_{\mathsf{upp}}(t) = C_{\mathsf{upp}}(t ; \cA, D) \geq 0$ such that
\bes{
\bbP_{A \sim \cA} \{ \nm{A(f)} \geq t \nm{f} \} \leq C_{\mathsf{upp}}(t ; \cA,D),\quad \forall f \in D.
}
Next, given $t,s \geq 0$ an \textit{(upper) absolute concentration bound for $\cA$} is any constant $C_{\mathsf{abs}}(s,t ; \cA , D)$ such that
\bes{
\bbP_{A \sim \cA}(\nm{A(f)} > t ) \leq C_{\mathsf{abs}}(s,t ; \cA , D),\quad \forall f \in D,\ \nm{f} \leq s.
 }
Similarly, \textit{(upper) concentration bound for $\cE$} is any constant $D_{\mathsf{upp}}(t) = D_{\mathsf{upp}}(t ; \cE) \geq 0$ such that
\bes{
\cE(B^c_t) = \bbP_{e \sim \cE}(\nm{e} \geq t)  \leq D_{\mathsf{upp}}(t ; \cE).
}
The four previous quantities are similar to those in \cite{adcock_how_2025} (the main difference is the extension to infinite-dimensional Hilbert spaces). We also introduce the following concept, which differs from that in \cite{adcock_how_2025}, and is better suited to the infinite-dimensional setting.
Given $\tau,\varepsilon \geq 0$, a \textit{density shift bound} for $\cM$ is any constant $D_{\mathsf{shift}}( \varepsilon,\tau) = D_{\mathsf{shift}}(\varepsilon,\tau ; \cM) \geq 0$ (possibly $+\infty$) such that
\bes{
\frac{\D \cE_A}{\D T_{v} \sharp \cE_A}(e) \leq D_{\mathsf{shift}}(\varepsilon,\tau ; \cM),\quad \forall e,v \in \bbR^m,\ \nm{e} \leq \tau,\ \nm{v} \leq \varepsilon, \quad \text{a.s.}\ A \sim \cA.
}
Here $\cE_A$ is the conditional distribution of $e$, given $A$, and $T_{v} : \bbX \rightarrow \bbX$, $u \mapsto u + v$ is the translation map. Note we set $D_{\mathsf{shift}}(\varepsilon,\tau) = + \infty$ if $\cE_A$ is not absolutely continuous with respect to the pushforward $T_{v} \sharp \cE_A$ for some $v$ and $A$. A probability measure for which $\cE_A$ is equivalent to $T_{v} \sharp \cE_A$ for all $v \in \bbR^m$ is called \textit{quasi-invariant}. Hence $D_{\mathsf{shift}}(\varepsilon) < \infty$ whenever $\cE_A$ is quasi-invariant \textit{and} the Radon-Nikodym derivative $\D \cE_A / \D T_v \sharp \cE_A$ is bounded almost surely for $A \sim \cA$.
This will be the case in this work whenever this constant is used.

\subsection{Abstract result}\label{ss:abstract-result}

We now establish an abstract result that bounds \ef{p-general-def} at a high level of generality, namely, it makes very few assumptions on the distributions $\cR$, $\cP$, $\cA$ and $\cE$. We do this with a series of lemmas that extend those in \cite{adcock_how_2025} to the Hilbert space setting.

\lem{[Separation lemma] Let $\cH_1, \ldots, \cH_k$ be Borel probability measures on $\bbX$ and consider the mixture $\cH = \sum_{i=1}^k a_i \cH_i$. Let $y^* \sim \cH$ and $\hat{y} \sim \sum_{i=1}^k \bbP(y^* \sim \cH_i |y^*) \cH_i (\cdot | y^*)$ where $\bbP(y^* \sim \cH_i|y^*)$ are the posterior weights. Then 
\bes{
    \bbP[y^* \sim \cH_i, \hat{y} \sim \cH_j (\cdot |y^*)] \leq 1 - \mathrm{TV}(\cH_i, \cH_j).
}
}
The proof of this lemma is identical to \cite[Lem.\ C.4]{adcock_how_2025}, as it does not make any assumptions on the domain of the probability measures.

\lem{
[Disjointly-supported measures induce well-separated measurement distributions]
\label{l:disjoint-support-TV-dist}
Let $\tilde{f} \in \bbX$, $\sigma \geq 0$, $\eta \geq 0$, $c \geq 1$, $\cP_{\mathsf{ext}}$ be a distribution supported in the set 
\begin{equation*}
    S_{\tilde{f}, \mathsf{ext}} = \{ f \in \bbX : \nm{f - \tilde{f}} \geq c(\eta + \sigma) \}
\end{equation*}
and $\cP_{\mathsf{int}}$ be a distribution supported in the set
\begin{equation*}
S_{\tilde{f}, \mathsf{int}} = \{ f \in \bbX : \nm{f - \tilde{f}} \leq \eta \}.
\end{equation*}
Given $A \in \cL(\bbX_0 , \bbR^m)$, let $\cH_{\mathsf{int}, A}$ be the distribution of $y = A(f^*) + e$ where $f^* \sim \cP_{\mathsf{int}}$ and $e \sim \cE_A$ independently, and define $\cH_{\mathsf{ext}, A}$ in a similar way.
Then 
\begin{equation*}
    \bbE_{A \sim \cA}[\mathrm{TV}(\cH_{\mathsf{int}, A}, \cH_{\mathsf{ext}, A})] \geq 1 - \left [ C_{\mathsf{low}}  \left ( \frac{2}{\sqrt{c}} ; \cA , D_{\mathsf{ext}} \right )  + C_{\mathsf{upp}} \left (\frac{\sqrt{c}}{2} ; \cA , D_{\mathsf{int}} \right ) + 2 D_{\mathsf{upp}} \left ( \frac{\sqrt{c} \sigma}{ 2} ; \cE \right ) \right ],
\end{equation*}
where $D_{\mathsf{ext}} =  \{ f - \tilde{f} : f \in\mathrm{supp}(\cP_{\mathsf{ext}}) \}$ and $D_{\mathsf{int}} =  \{ f - \tilde{f} : f \in\mathrm{supp}(\cP_{\mathsf{int}}) \}$.
}

\prf{
The proof is similar to that of \cite[Lem.\ C.5]{adcock_how_2025}, except with the changes that come from working in general Hilbert spaces and not assuming independence of $A$ and $e$. We detail these changes.

We first define the set
 \bes{
B_A = \{ y \in \bbR^m : \nm{y - A\tilde{f}} \leq \sqrt{c} (\eta + \sigma)\}
}
and observe that
\be{
\label{TV-split}
\bbE_{A \sim \cA}[\mathrm{TV}(\cH_{\mathsf{int}, A}, \cH_{\mathsf{ext}, A})] \geq \bbE_{A \sim \cA} [\cH_{\mathsf{int}, A}(B_A)] - \bbE_{A \sim \cA}[\cH_{\mathsf{ext}, A}(B_A)].
}
Consider the first term. We write
\bes{
\bbE_{A \sim \cA} [\cH_{\mathsf{ext}, A}(B_A)] = I_1 + I_2,
}
Given $f \in \bbX$, let $C_f = \{ A : \nm{A(f) - A(\tilde{f})} < 2 \sqrt{c}(\eta + \sigma) \} \subseteq \cL(\bbX_0, \bbR^m)$ and write
\bes{
I_{1} = \bbE_{f \sim \cP_{\sfext}} [\bbE_{A \sim \cA} \cE_A(B_A - A(f))1_{C_f}], \quad I_{2} = \bbE_{f \sim \cP_{\sfext}} [\bbE_{A \sim \cA} \cE_A(B_A - A(f))1_{C_f^c}]. 
}
Identically to \cite[Lem.\ C.5]{adcock_how_2025}, we have $I_1 \leq C_{\mathsf{low}}(2/\sqrt{c} ; \cA , D_{\mathsf{ext}})$. For $I_2$, we write
\bes{
I_2 \leq \bbE_{f \sim \cP_{\mathsf{ext}}}\left [ \bbE_{A \sim \cA} [ \cE(B_A - A(f) | A) \right ].
}
Observe that $B_A - A(f) \subseteq B^c_{\sqrt{c}(\eta+\sigma)}$. Hence, by the tower property,
\bes{
I_2 \leq \bbE_{f \sim \cP_{\mathsf{ext}}} \cE(B^c_{\sqrt{c}(\eta+\sigma)}) \leq D_{\mathsf{upp}}(\sigma \sqrt{c} ; \cE).
}
We deduce that
\bes{
\bbE_{A \sim \cA} [\cH_{\mathsf{ext}, A}(B_A)] \leq C_{\mathsf{low}} \left (\frac{2}{\sqrt{c}} ; \cA , D_{\mathsf{ext}} \right ) + D_{\mathsf{upp}}\left(\frac{\sqrt{c}}{2}\sigma  ; \cE \right )
}
The argument for the other term of \ef{TV-split} follows similar arguments. We omit the details.
}

\lem{
[Replacing the real distribution with the approximate distribution]
\label{lem:measure-replacement}
Let $\varepsilon , \sigma , d, t \geq 0$, $c \geq 1$, $\Pi$ be an $W_{\infty}$-optimal coupling of $\cR$ and $\cP$ and define the set $D = \{ f^* - g^* : (f^*,g^*) \in \mathrm{supp}(\Pi) \}$.
Let
 \bes{
p = \bbP_{f^* \sim \cR, (A, e) \sim \cM,\hat{f} \sim \cP(\cdot | A (f^*)+e , A)} [ \nm{f^* - \hat{f}} \geq d + \varepsilon]
 }
 and
 \bes{
q = \bbP_{g^* \sim \cP, (A, e) \sim \cM,\hat{g} \sim \cP(\cdot | A (g^*)+e , A)}  [\nm{g^*  - \hat{g}} \geq d].
 }
 Then
\bes{
p \leq C_{\mathsf{abs}}(\varepsilon , t \varepsilon ; \cA, D ) + D_{\mathsf{upp}}(c \sigma ; \cE) + D_{\mathsf{shift}}(t \varepsilon , c \sigma ; \cM) q .
}
}
\prf{
Similarly to the previous proofs, the arguments are broadly similar to those of \cite[Lem.\ C.6]{adcock_how_2025}, with the necessary changes to account for the more general setup.
Define the events
\bes{
B_{1, \hat{f}} = \{ f^* : \nm{f^* - \hat{f}} \geq d + \varepsilon\},\quad  B_{2, \hat{f}} = \{ g^* : \nm{\hat{g} - g^*} \geq d \}
}
so that 
    \begin{align*}
        p &= \bbP_{f^* \sim \cR, (A, e) \sim \cM, \hat{f} \sim \cP(\cdot | A(f^*) + e, A)} [f^* \in B_{1, \hat{f}}] \\
        q &= \bbP_{g^* \sim \cP, (A, e) \sim \cM, \hat{g} \sim \cP(\cdot | A(g^*) + e, A)} [g^* \in B_{2, \hat{g}}].
    \end{align*}
With the assumption that $W_\infty (\cR, \cP) \leq \varepsilon$, there exists a coupling such that $\Pi(\nm{f^* - g^*} \leq \varepsilon) = 1$. This gives
\begin{align*}
      p  =  \int \int \int \int 1_{B_{1, \hat{f}}}(f^*) \D \cP(\cdot | A(f^*) + e, A)(\hat{f}) \D \cM(A, e) \D \Pi(f^*, g^*).
\end{align*}
Define $E = \{(f^*, g^*) : \nm{f^* - g^*} \leq \varepsilon \}$ and observe $\Pi(E) = 1$. Then for fixed $A, e$, and $(f^*, g^*) \in E$, we have 
\begin{equation*}
        \int 1_{B_{1, \hat{f}}}(f^*) \D \cP(\cdot | A(f^*) + e, A)(\hat{f}) \leq \int 1_{B_{2, \hat{f}}}(g^*) \D \cP(\cdot | A(f^*) + e, A)(\hat{f}).
\end{equation*}
By Fubini's theorem, which we note we can invoke by non-negativity of indicator functions, we get
\eas{
p & = \int \int \int \int 1_{B_{1, \hat{f}}}(f^*) \D \cP(\cdot | A(f^*) + e, A)(\hat{f}) \D \cM(A, e) \D \Pi(f^*, g^*) 
\\
&\leq \int \int \int \int 1_{B_{2, \hat{f}}}(g^*) \D \cP(\cdot | A(f^*) + e, A)(\hat{f}) \D \cM(A, e) \D \Pi(f^*, g^*).
}
Define $C_{f^*, g^*} = \{A : \nm{A(f^* - g^*)} > t \varepsilon\}$ and 
\begin{align*}
        I_1 & = \int \int 1_{C_{f^*, g^*}}(A) \int \int 1_{B_{2, \hat{f}}}(g^*) \D \cP(\cdot | A(f^*) + e, A)(\hat{f}) \D \cM(A, e) \D \Pi(f^*, g^*)  \\
        I_2 & = \int \int 1_{C_{f^*, g^*}^c}(A) \int \int 1_{B_{2, \hat{f}}}(g^*) \D \cP(\cdot | A(f^*) + e, A)(\hat{f}) \D \cM(A, e) \D \Pi(f^*, g^*) 
    \end{align*}
so that 
\bes{
    p \leq I_1 + I_2.
}
The bound for $I_1$ goes identically to \cite[Lem \, C.3]{adcock_how_2025} and gives $I_1 \leq C_{\mathsf{abs}}(\varepsilon,t \varepsilon ; \cA , D)$. To bound $I_2$, we further split the integral as follows: we write $I_2 = I_{2_1} + I_{2_2}$, where
\eas{  
 I_{2_1} = & \int \int 1_{C_{f^*, g^*}^c}(A) \int 1_{B^c_{c\sigma}}(e)\int 1_{B_{2, \hat{f}}}(g^*) \D \cP(\cdot | A(f^*) + e, A)(\hat{f}) \D \cM(A, e) \D \Pi(f^*, g^*) \\
        \\
I_{2_2} = & \int \int 1_{C_{f^*, g^*}^c}(A) \int 1_{B_{c\sigma}}(e)\int 1_{B_{2, \hat{f}}}(g^*) \D \cP(\cdot | A(f^*) + e, A)(\hat{f}) \D \cM(A, e) \D \Pi(f^*, g^*)
}
To find a bound for $I_{2_1}$, we use a similar argument to Lemma \ref{l:disjoint-support-TV-dist}.  We have
\eas{
    I_{2_1} & \leq \int \int 1_{C_{f^*, g^*}^c}(A) \int 1_{B^c_{c\sigma}}(e)  \D \cE_A(e) \D \cA(A) \D \Pi(f^*, g^*) & \leq  \int \int 1_{B^c_{c\sigma}}(e) \D \cE_A(e) \D \cA(A),
}
and therefore, by the tower property, 
\bes{
        I_{2_1} = \bbE_{A \sim \cA}[\cE_A(B_{c\sigma}^c)] = \cE(B_{c\sigma}^c) \leq D_{\mathsf{upp}}(c\sigma; \cE).
    }
    Now consider $I_{2_2}$. For fixed $A \in C^c_{f^*,g^*}$, define the random variable $e' = e + A(f^* - g^*)$. Suppose, without loss of generality, that $D_{\mathsf{shift}}(t \varepsilon , c \sigma ; \cM) < + \infty$. Then the Radon-Nikodym derivative $\D \cE_A / \D \cE'_A(e)$ exists and is finite for $e \in B_{c \sigma}$, since $v = A(f^* - g^*) = e - e'$ satisfies $\nm{v} \leq t \varepsilon$. Therefore
    \eas{
      I_{2_2}    \leq &~ D_{\mathsf{shift}}(t\varepsilon, c\sigma; \cM) \int \int 1_{C_{f^*, g^*}^c}(A) \int 1_{B_{c\sigma}}(\omega) \\
         & \times \int 1_{B_{2, \hat{f}}}(g^*) d\cP(\cdot | A(f^*) + e, A)(\hat{f}) \D \cE_{A}'(e)  \D \cA(A)  \D \Pi(f^*, g^*).
    }
We now change variables back with $e = e' - A(f^* - g^*)$ to get
    \eas{
        I_{2_2}  = &~ D_{\mathsf{shift}}(t\varepsilon, c\sigma; \cM) \int \int 1_{C_{f^*, g^*}^c}(A) \int 1_{B_{c\sigma}}(e' - A(f^* - g^*))\\
         & \times \int 1_{B_{2, \hat{f}}}(g^*) d\cP(\cdot | A(g^*) + e', A)(\hat{f}) \D \cE_A(e')  \D \cA(A)  \D \Pi(f^*, g^*).
    }
Next we replace the variables $\hat{f}$ and $e'$ with $\hat{g}$ and $e$ respectively to get
    \eas{
        I_{2_2}  =&  D_{\mathsf{shift}}(t\varepsilon, c\sigma; \cM) \int \int 1_{C_{f^*, g^*}^c}(A) \int 1_{B_{c\sigma}}(e - A(f^* - g^*))\\
         ~ & \times \int 1_{B_{2, \hat{g}}}(g^*) d\cP(\cdot | A(g^*) + \omega, A)(\hat{g}) \D \cE_A(e)  \D \cA(A)  \D \Pi(f^*, g^*)
    }
and finally, we replace the inner term with $q$ to get
    \bes{
I_{2_2} \leq D_{\mathsf{shift}}(t\varepsilon, c\sigma; \cM) q.
    }
Combining with the estimates for $I_1$ and $I_{2_1}$ yields the result.
}

The following lemma is virtually identical to \cite[Lem.\ C.7]{adcock_how_2025}, the only difference being that the distributions now take values in $\bbX$. The proof is identical, as it only requires $\bbX$ to be a Polish space, which holds by assumption as $\bbX$ is a separable Hilbert space.

\lem{
[Decomposing distributions]
\label{lem:Coupling-lemma}
Let $\cR, \cP$ be arbitrary distributions on a separable Hilbert space $\bbX$, $p \geq 1$ and $\eta, \rho, \delta > 0$. 
If $W_p (\cR, \cP) \leq \rho$ and $k \in \bbN$ is such that
\be{
\label{cover-numb-min-k}
\min \{ \log \mathrm{Cov}_{\eta, \delta}(\cP), \log \mathrm{Cov}_{\eta, \delta}(\cR) \} \leq k, 
}
then there exist distributions $\cR', \cR'', \cP', \cP''$, a constant $0 < \delta' \leq \delta$ and a  discrete distribution $\cQ$ with $\mathrm{supp}(\cQ) = S$ satisfying
\begin{enumerate}
    \item $\min \{ W_\infty (\cP', \cQ), W_\infty(\cR', \cQ) \} \leq \eta$,
    \item $W_\infty (\cR', \cP') \leq \frac{\rho}{\delta^{1/p}}$,
    \item $\cP = (1 - 2\delta') \cP' + (2\delta')\cP''$ and $\cR = (1 - 2\delta')\cR' + (2\delta')\cR''$,
    \item $|S| \leq \E^{k}$,
    \item and $S \subseteq \mathrm{supp}(\cP)$ if $\cP$ attains the minimum in \ef{cover-numb-min-k} with $S \subseteq \mathrm{supp}(\cR)$ otherwise.
\end{enumerate}
}

We can now establish the following abstract result, which extends \cite[Thm.\ 3.1]{adcock_how_2025} to the Hilbert spaces.

\thm{
\label{t:main-res-abstract}
Let $1 \leq p \leq \infty$, $0 \leq \delta \leq 1/4$, $\varepsilon, \eta, t > 0$, $c ,c' \geq 1$ and $\sigma \geq \varepsilon / \delta^{1/p}$. Let $\cE$ be a distribution on $\bbR^m$ and $\cR, \cP$ be distributions in $\bbX$ satisfying $W_p(\cR, \cP) \leq \varepsilon$ and 
\be{
\label{min-cov-k-main}
\min (\log \mathrm{Cov}_{\eta, \delta} (\cR), \log \mathrm{Cov}_{\eta, \delta} (\cP)) \leq k
}
for some $k \in \bbN$. Suppose that $f^* \sim \cR$, $(A, e) \sim \cM$ independently and $\hat{f} \sim \cP(\cdot | y , A)$, where $y = A (f^*) + e$. 
Then $p : = \bbP[\nm{f^* - \hat{f}} \geq (c+2)(\eta + \sigma)] $ satisfies
\eas{
& p
 \leq  2 \delta +  C_{\mathsf{abs}}(\varepsilon / \delta^{1/p} , t \varepsilon / \delta^{1/p} ; \cA , D_1) + D_{\mathsf{upp}}(c'\sigma ; \cE) 
\\
& + 2 D_{\mathsf{shift}}(t \varepsilon / \delta^{1/p} , c' \sigma ; \cM) \E^k  \Bigg [ C_{\mathsf{low}} \left ( \frac{2 \sqrt{2}}{\sqrt{c}} ; \cA , D_2 \right ) + C_{\mathsf{upp}} \left ( \frac{\sqrt{c}}{2\sqrt{2}} ; \cA , D_2 \right ) + 2 D_{\mathsf{upp}} \left ( \frac{\sqrt{c} \sigma}{2 \sqrt{2}} ; \cE \right ) \Bigg ],
}
where
\be{
\label{D1-set-main}
D_1 = B_{\varepsilon / \delta^{1/p}}(\mathrm{supp}(\cP)) \cap \mathrm{supp}(\cR) - \mathrm{supp}(\cP)
}
and
\be{
\label{D-set-main}
D_2  = \begin{cases}\mathrm{supp}(\cP) - \mathrm{supp}(\cP) & \text{if $\cP$ attains the minimum in \ef{min-cov-k-main}} \\ \mathrm{supp}(\cR) - \mathrm{supp}(\cP) & \text{otherwise} \end{cases}.
}
}
\prf{
The proof is virtually identical, except for a couple of changes. First, throughout the proof, we replace $A\sim \cA$, $e \sim \cE$ with $(A,e) \sim \cM$ to account for the joint model considered in this work. Second, before \cite[Eqn.\ (C.23)]{adcock_how_2025}, we fix $A \in \cL(\bbX_0,\bbR^m)$ and replace $e \sim \cE$ with $e \sim \cE_A$.
}

\subsection{Proof of Theorem \ref{t:sensor-distribution}}

The proof of Theorem \ref{t:sensor-distribution} is based on Theorem \ref{t:main-res-abstract}, after making a suitable choice for the distribution $\cM$ and estimating the various constants. Specifically, consider the setup of \S \ref{s:theory}, suppose that the sampling distribution $\mu$ satisfies the conditions in Theorem \ref{t:sensor-distribution}, i.e., $\mu \ll \rho$, $ \D \mu / \D \rho > 0$ $\mu$-a.s.\ and let $w(\theta) = \left ( \D \mu(\theta) / \D \rho \right )^{-1}$. Then we define $(A,e) \sim \cM$ as
\bes{
A(f) = \left ( \sqrt{\frac{w(\theta_i)}{m}} L(\theta_i)(f) \right )^{m}_{i=1},\quad \forall f \in \bbX_0
}
and
\bes{
e = \left (  \sqrt{\frac{w(\theta_i)}{m}} n_i \right )^{m}_{i=1},
}
where $\theta = (\theta_i)^{m}_{i=1} \sim \mu^{\otimes m}$ and $n \sim \cN(0,\sigma^2 I)$ independently. Given $f^* \in \bbX_0$, we now consider its recovery via posterior sampling $\cP(\cdot | b,A)$, where $b = A(f^*) + e$. 
Note that this posterior is identical to that considered in Theorem \ref{t:sensor-distribution}, i.e., $\cP(\cdot | b , A) = \cP(\cdot | y , \theta)$ with $\theta$ and $y$ given by \ef{data-model}. This can be shown as follows. Notice first that $b = W(\theta) y$ and $A = W(\theta) L(\theta)$, where $W(\theta) = \mathrm{diag}(\sqrt{w(\theta_1)/m},\ldots,\sqrt{w(\theta_m)/m} )$. Recall that $w > 0$ a.s., therefore $W(\theta)$ is measurable and invertible a.s.. Moreover, the map $(y,\theta) \mapsto (b,A)$ is injective, since $\theta \in \Theta \mapsto L(\theta) \in \cB(\bbX_0)$ is injective by assumption. The claim now follows.
Hence, for the remainder of the proof, we may consider $\cP(\cdot | b,A)$. 

\rem{
\label{rem:dependence}
A critical observation is that the sensing operator $A$ and noise $e$ are now dependent random variables, due to the weighting $w(\theta_i)$. Note that this weighting is essential, as it ensures that
\bes{
\bbE \nm{A(f)}^2 = \sum^{m}_{i=1} \bbE \frac{w(\theta_i)}{m} | L(\theta_i)(f) |^2 = \int_{\Theta} w(\theta) | L(\theta)(f) |^2 \D \mu(\theta) = \int_{\Theta} |L(\theta)(f) |^2 \D \rho(\theta).
}
Therefore, by \ef{nondegeneracy}, $\bbE \nm{A(f)}^2$ is equivalent, up to $\alpha$ and $\beta$, to $\nm{f}^2$. This crucial observation allows us to derive concentration bounds $C_{\mathsf{low}}$ and $C_{\mathsf{upp}}$ that are exponentially-small in $m$ (Lemma \ref{l:rel-conc-bound-A}), which is a key step in proving Theorem \ref{t:sensor-distribution}.

The  dependence between $A$ and $e$ means, in particular, that the framework of \cite{adcock_how_2025} cannot be used, as it assumes independence between these two terms. Fortunately, Theorem \ref{t:main-res-abstract} extends \cite{adcock_how_2025} by allowing dependence and considering only their joint distribution $\cM$. This key generalization facilitates OSP and the proof of Theorem \ref{t:sensor-distribution}.
}

Having defined the distribution $\cM$, the next step is to estimate the various constants appearing in Theorem \ref{t:sensor-distribution}.

\lem{[Relative concentration bounds for $\cA$]
\label{l:rel-conc-bound-A}
Let $D  = \bbS(\cP)$ and suppose that \ef{nondegeneracy} holds with constants $0 < \alpha \leq \beta < \infty$. Then the relative lower and upper concentration bounds for $\cA$ can be taken as
\eas{
C_{\mathsf{upp}}(t ; \cA , D ) &= 2 \exp \left ( - \frac{m c_{\mathsf{upp}}(t,\beta)}{\kappa_w(\cP,L)} \right),\quad \forall t > \sqrt{\beta},
\\
C_{\mathsf{low}}(1/t ; \cA , D ) &= 2 \exp \left ( - \frac{m c_{\mathsf{low}}(t,\alpha)}{\kappa_w(\cP,L)} \right),\quad \forall t > 1/\sqrt{\alpha},
}
where $c_{\mathsf{upp}}(t,\beta)$ depends on $t$ and $\beta$ only, $c_{\mathsf{low}}(t,\alpha)$ depends on $t$ and $\alpha$ only and $\kappa_w(\cP,L)$ is as in \ef{kappa_w_def}.
}
\prf{
Let $f \in D$ be arbitrary and write
\bes{
\nm{A(f)}^2 = \frac1m \sum^{m}_{i=1} w(\theta_i) | L(\theta_i)(f)|^2 = \sum^{m}_{i=1} X_i,
}
where the $X_i$ are i.i.d.\ random variables with $X_i \geq 0$. Notice that
\bes{
\sum^{m}_{i=1} \bbE(X_i) = \int_{\Theta} w(\theta) |L(\theta)(f)|^2 \D \mu(\theta) = \int_{\Theta} | L(\theta)(f) |^2 \D \rho(\theta)
}
and therefore
\bes{
\alpha \nm{f}^2 \leq \sum^{m}_{i=1} \bbE(X_i) \leq \beta \nm{f}^2
}
by the nondegeneracy assumption \ef{nondegeneracy}. Hence
\bes{
\bbP(\nm{A(f)} \geq t \nm{f}) \leq \bbP \left ( \sum^{m}_{i=1} X_i \geq \frac{t^2}{\beta} \sum^{m}_{i=1} \bbE(X_i) \right )
}
and
\bes{
\bbP(\nm{A(f)} \leq t^{-1} \nm{f}) \leq \bbP \left ( \sum^{m}_{i=1} X_i \leq \frac{1}{t^2 \alpha} \sum^{m}_{i=1} \bbE(X_i) \right ).
}
Now observe that
\bes{
X_i \leq \kappa_w(\cP,L) / m
}
We now recall the standard Chernoff bound, which gives
\bes{
\bbP \left ( \sum^{m}_{i=1} X_i \geq (1+\delta) \sum^{m}_{i=1} \bbE(X_i) \right ) \leq \exp \left ( - \frac{m c_{\mathsf{upp}}(\delta) }{\kappa_w(\cP,L)} \right ),\quad \forall \delta > 0,
}
and
\bes{
\bbP \left ( \sum^{m}_{i=1} X_i \leq (1-\delta) \sum^{m}_{i=1} \bbE(X_i) \right ) \leq \exp \left ( - \frac{m c_{\mathsf{low}}(\delta) }{\kappa_w(\cP,L)} \right ),\quad \forall 0 < \delta < 1 .
}
Setting $\delta = t^2/\beta - 1$ and $\delta = 1-1/(\alpha t^2)$ now gives the result.
}

\lem{[Absolute concentration bound for $\cA$]
\label{l:abs-conc-bound-A}
Let $D = \bbX_0$ and suppose that \ef{nondegeneracy} holds with constants $0 < \alpha \leq \beta < \infty$. Then the absolute concentration bound for $\cA$ can be taken as
\bes{
C_{\mathsf{abs}}(s,t ; \cA , D) = \frac{s^2}{t^2}.
}
}
\prf{
By Markov's inequality, the definition of $A$ and \ef{nondegeneracy}, we have
\bes{
\bbP_{A \sim \cA} (\nm{A(f)} > t ) \leq \frac{\bbE \nm{A(f)}^2}{t^2} \leq \frac{\beta \nm{f}^2}{t^2}.
}
Therefore, if $\nm{f} \leq s$, we obtain $\bbP_{A \sim \cA} (\nm{A(f)} > t ) \leq s^2 / t^2$, as required.
}

\lem{
[Concentration bound for $\cE$] 
\label{l:conc-bound-E}
The upper concentration bound $D_{\mathsf{upp}}(t  ; \cE)$ can be taken as
\bes{
D_{\mathsf{upp}}(t ; \cE) = \left ( \frac{t^2}{w_{\max} \sigma^2} \E^{1-\frac{t^2}{w_{\max}\sigma^2}} \right )^{\frac{m}{2}},\quad \forall t > \sigma \sqrt{w_{\max}},
}
where $w_{\max} = \esssup_{\theta \sim \rho} w(\theta)$.
}
\prf{
Recall that $\cE$ is the marginal distribution, i.e., the distribution of $e$ when $(A,e) \sim \cM$. Observe that
\bes{
\cE(B^c_t) = \bbE_{A \sim \cA}[\bbP_{e \sim \cE_A}(\nm{e} > t)].
}
Now recall that $\cE_A  = \cN(0 , \frac{\sigma^2}{m} \diag (w(\theta_1),\ldots,w(\theta_m) ) )$. Hence
\eas{
\bbP_{e \sim \cE_A}(\nm{e} > t) &= \bbP_{n \sim \cN(0,\frac{\sigma^2}{m} I)} \left ( \sum^{m}_{i=1} w(\theta_i) n^2_i > t^2 \right ) 
\leq \bbP_{n \sim \cN(0,\frac{\sigma^2}{m} I)} \left ( \nm{n}^2 > \frac{t^2}{w_{\max}} \right ).
}
The result now follows from \cite[Lem.\ B.1]{adcock_how_2025}.
}

\lem{
[Density shift bound for $\cM$]
\label{l:dens-shifty-bd}
The density shift bound $D_{\mathsf{shift}}(\varepsilon,\tau ; \cM)$ can be taken as 
\bes{
D_{\mathsf{shift}}(\varepsilon,\tau ; \cM) = \exp \left ( \frac{m(2 \tau + \varepsilon)}{2 \sigma^2 w_{\min} } \varepsilon \right ) ,\quad \forall \varepsilon,\tau \geq 0,
}
where $w_{\min} = \mathrm{essinf}_{\theta \sim \rho} w(\theta)$.
}
\prf{
Given $\theta$, let $A$ be as defined above and $e,v \in \bbR^m$ with $\nm{e} \leq \tau$ and $\nm{v} \leq \varepsilon$.
Recall from the previous proof that $\cE_A  = \cN(0 , \frac{\sigma^2}{m} \diag (w(\theta_1),\ldots,w(\theta_m) ) )$ and therefore $T_v \sharp \cE_A  = \cN(v , \frac{\sigma^2}{m} \diag (w(\theta_1),\ldots,w(\theta_m) ) )$. Both distributions are absolutely continuous with respect to Lebesgue measure. Hence the Radon-Nikodym derivative $\D \cE_A / \D T_v \sharp \cE_A$ is just the ratio of their densities, i.e.,
\eas{
\frac{\D \cE_A}{\D T_v \sharp \cE_A}(e) &= \exp \left [ \frac{m}{2 \sigma^2} \sum^{m}_{i=1} \left ( \frac{(e_i - v_i)^2}{w(\theta_i)}  - \frac{e^2_i}{w(\theta_i)} \right ) \right ] = \exp \left [ \frac{m}{2 \sigma^2} \sum^{m}_{i=1} \frac{v_i (v_i - 2 e_i)}{w(\theta_i)}  \right ] .
}
By the Cauchy--Schwarz inequality and the assumptions on $v$ and $e$, we have
\bes{
\sum^{m}_{i=1} \frac{v_i (v_i - 2 e_i)}{w(\theta_i)}  \leq \frac{\nm{v} \nm{v - 2 e}}{w_{\min}}  \leq \frac{(\varepsilon + 2 \tau) \varepsilon}{w_{\min}}
}
The result now follows immediately.
}

\prf{[Proof of Theorem \ref{t:sensor-distribution}]
Let $p = \bbP [ \nm{f^* - \hat{f} } \geq (8 d^2+2) (\eta+\sigma)]$ for some $d \geq 1$ to be chosen later. Now apply Theorem \ref{t:main-res-abstract} with $c = 8 d^2$ to get
\eas{
p \lesssim &~ \delta + C_{\mathsf{abs}}(\tilde{\varepsilon} , t \tilde{\varepsilon} ; \cA , \bbX_0) + D_{\mathsf{upp}}(c' \sigma ; \cE) 
\\
& + D_{\mathsf{shift}}(t \tilde{\varepsilon} , c' \sigma  ; \cM) \exp(k) \left [ C_{\mathsf{low}} \left ( \frac{1}{d} ; \cA , D \right ) + C_{\mathsf{upp}} \left ( d ; \cA , D \right ) + D_{\mathsf{upp}} \left ( d \sigma ; \cE \right ) \right ]
}
for any $c' \geq 1$ and $t > 0$, where $\tilde{\varepsilon} = \varepsilon / \delta^{1/p}$, $D = \mathrm{supp}(\cP) - \mathrm{supp}(\cP)$ and $k \in \bbN$ is any integer such that
\bes{
\log \mathrm{Cov}_{\eta,\delta}(\cP) \leq k.
}
Set
\bes{
c' = 2 \sqrt{w_{\max}},\quad  d = 2 \max \{ \sqrt{w_{\max}} , 1/\sqrt{\alpha} , \sqrt{\beta} \}.
}
Then Lemma \ref{l:conc-bound-E} gives
\bes{
D_{\mathsf{upp}} \left ( d \sigma ; \cE \right ) \leq  D_{\mathsf{upp}} \left ( c' \sigma ; \cE \right ) \leq  \left ( 2 \E^{-1} \right )^{\frac{m}{2}} \leq \exp(-m/16).
}
Now set $t = 1/\sqrt{\delta}$. Then Lemma \ref{l:abs-conc-bound-A} gives
\bes{
 C_{\mathsf{abs}}(\tilde{\varepsilon} , t \tilde{\varepsilon} ; \cA , \bbX_0) = \frac{\tilde{\varepsilon}^2}{t^2 \tilde{\varepsilon}^2} = \delta.
}
Next, Lemma \ref{l:rel-conc-bound-A} and the choice of $d$ gives
\bes{
C_{\mathsf{low}} \left ( \frac1d ; \cA , D \right ) + C_{\mathsf{upp}} \left ( d ; \cA , D \right ) \leq 2 \exp \left ( - \frac{m c(\alpha,\beta)}{\kappa_w(\cP)} \right ),
}
for some constant $c(\alpha,\beta)$ depending on $\alpha,\beta$ only.
Substituting these three bounds into the previous expression, we deduce that
\eas{
p \lesssim  & ~ \delta + \exp\left(-\frac{m}{16} \right) 
 + D_{\mathsf{shift}}( t \tilde{\varepsilon} , c' \sigma ; \cM) \exp(k) \left [ \exp \left ( - \frac{m c(\alpha,\beta)}{\kappa_w(\cP)} \right ) + \exp\left(-\frac{m}{16} \right) \right ].
}
Finally, we estimate the density shift bound. Using Lemma \ref{l:dens-shifty-bd}, we get
\eas{
D_{\mathsf{shift}}( t \tilde{\varepsilon} , c' \sigma ; \cM) \leq \exp \left ( \frac{m(2 c' \sigma + t \tilde{\varepsilon})}{2 \sigma^2 w_{\min}} t \tilde{\varepsilon} \right )
 = \exp \left ( \frac{2 m \sqrt{w_{\max}} \tilde{\varepsilon}}{\sigma w_{\min} \sqrt{\delta}} + \frac{m \tilde{\varepsilon}^2}{2 \sigma^2 w_{\min} \delta } \right ).
}
Now suppose that
\be{
\label{tilde-eps-cond}
\tilde{\varepsilon} \leq \frac{w_{\min} \sqrt{\delta}}{m} \sigma.
}
Notice that this choice satisfies $\tilde{\varepsilon} \leq \sigma$, since $\delta \leq 1$ by assumption and $w_{\min} \leq 1$. The latter observation follows from the fact that $1 = \int_{D} \D \mu(\theta) = \int w(\theta)^{-1} \D \rho(\theta)$. This yields
\bes{
D_{\mathsf{shift}}( t \tilde{\varepsilon} , c' \sigma ; \cM) \leq \exp (2 \sqrt{w_{\max}} + 1 ) \leq \exp(3 \sqrt{w_{\max}}),
}
where in the final step we use the observation that $w_{\max} \geq 1$, which follows from the same fact as was used directly above. To conclude, we have shown that
\bes{
 \bbP [ \nm{f^* - \hat{f} } \geq (8 d^2+2) (\eta+\sigma)] \lesssim \delta + \exp \left ( 3 \sqrt{w_{\max}} + k - m \min \left \{ \frac{ c(\alpha,\beta)}{\kappa_w(\cP)} , \frac{1}{16} \right \} \right ) ,
}
provided that \ef{tilde-eps-cond} holds, where $ d = 2 \max \{ \sqrt{w_{\max}} , 1/\sqrt{\alpha} , \sqrt{\beta} \}$. Since $\tilde{\varepsilon} = \varepsilon / \delta^{1/p}$, we see that \ef{tilde-eps-cond} is equivalent to
\bes{
\sigma \geq \frac{m \varepsilon}{w_{\min} \delta^{1/p+1/2}},
}
which holds by assumption. To complete the proof, we use \ef{Wp-sigma-cond}.
}

\section{Discussion on the constant $C(\cP, L)$} \label{app:christoffel-dicussion}

Corollary \ref{cor:christoffel-sampling} shows the number of measurements that suffice for accurate recovery via Christoffel sampling is proportional to the intrinsic complexity of $\cP$, measured by $\log \mathrm{Cov}_{\eta,\delta}(\cP)$, up to the constant $C(\cP,L)$. In general, it is difficult to find an explicit expression or upper bound for the constant $C(\cP,L)$, as it depends critically on both $\cP$ and $L$. An exception is the case where $\cP$ is supported in certain intrinsically low-dimensional manifolds. For example, suppose that $\cP$ is supported in a $k$-dimensional subspace of $\bbX_0$. Let $\{ \phi_i \}^{k}_{i=1}$ be an orthonormal basis for this subspace and observe that, in this case, the generalized Christoffel function $K(\cP,L)$ can be expressed as
$$
K(\cP,L) = \sup_{\substack{c = (c_i)^{k}_{i=1} \in \bbC^d \\ \nm{c} = 1}} \left | L(\theta)\left(\sum^{d}_{i=1} c_i \phi_i \right ) \right |^2
$$
By linearity of $L(\theta)(\cdot)$ and the Cauchy--Schwarz inequality, we deduce that
$$
K(\cP,L) \leq \sum^{d}_{i=1} | L(\theta)(\phi_i) |^2.
$$
Using this, the nondegeneracy condition \ef{nondegeneracy} and the fact that the $\phi_i$ have unit $L^2_{\rho}$-norms, we deduce that
\bes{
C(\cP,L) = \int_{\Theta} K(\cP,L)(\theta) \D \rho(\theta) \leq \sum^{k}_{i=1} \int_{\Theta} | L(\theta)(\phi_i) |^2 \D \rho(\theta) \leq \beta k.
}
Thus, $C(\cP,L) \leq \beta k$ is at most linear in $k$, the dimension of the subspace in which $\mathrm{supp}(\cP)$ is contained.
More generally, if $\cP$ is supported in a union of $d$ subspaces of dimension at most $k$, then one has $C(\cP,L) \leq \beta k d$. This illustrates several scenarios in which $C(\cP,L)$ is controllable, even for arbitrary sensing operators $L$. More generally, while it may not be possible to analytically estimate $C(\cP,L)$ for complicated priors $\cP$, one can always estimate it numerically by computing the integral of the Christoffel function.

\section{Experimental details}\label{app:experiments}

In this appendix, we provide further details for the three main experiments shown in the paper. Here, we disclose that LLMs were used to assist in the coding of these experiments. All experiments and code however, has been checked and verified to the best of our ability to be accurate and not misleading.

\subsection{Pinball problem}\label{app:experiments-pinball}

\begin{figure}[tbp]
  \centering
  \includegraphics[width=\linewidth]{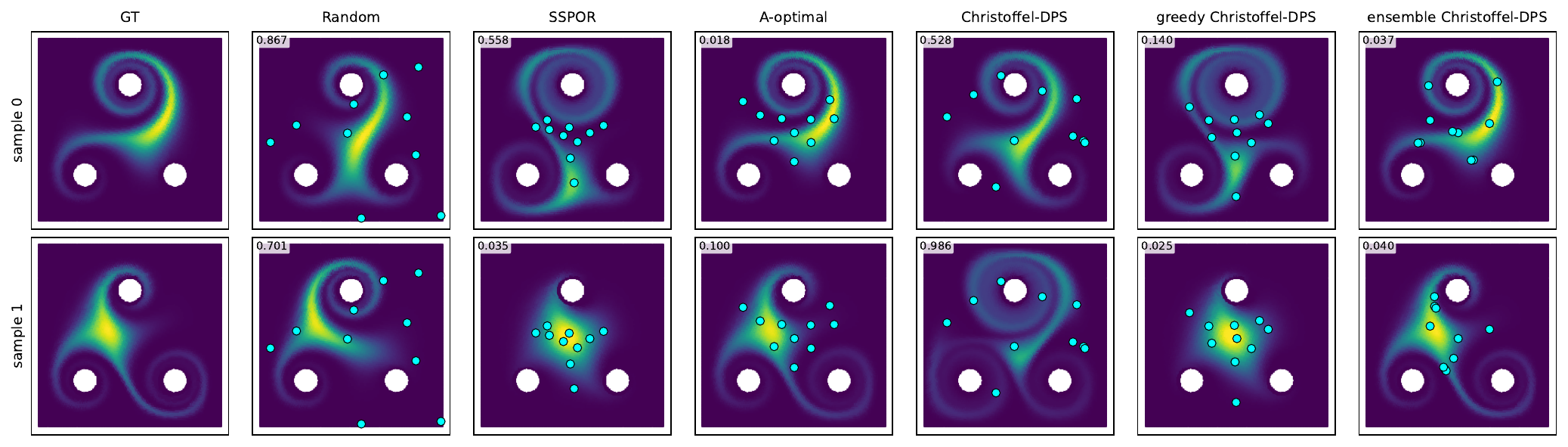}
   \caption{\textbf{Pinball reconstructions with $m{=}10$:} sensor locations on the unstructured $7525$-node mesh. Sensor positions overlaid; reconstruction relative L2 error annotated.}
  \label{fig:pinball_recon_k10}
\end{figure}

As a primary example, we consider the pinball dataset found in \cite{tomasetto2025reduced}. The setup and details for this experiment are very similar to \cite{rowbottom_grifdir_2026}. We restate them here for clarity.  Mathematically, this dataset consists of samples from an advection diffusion problem with an implicit parametric dependence on the square domain $[-1, 1] \times [-1, 1]$ with three cylinders of radius $r = 0.15$ at points $(-0.5, -0.5), (0.5, -0.5)$ and $(0.0, 0.5)$. We denote this square domain with the cylinders removed as $\Theta$. Each cylinder rotates at a constant velocity $v_i$, which we parameterize by $\mu = [v_1, v_2, v_3]$. The rotation of the cylinders with velocity $\mu$ induces a velocity $v: \Theta \to \bbR$ and a pressure $p : \Theta \to \bbR$ at each point in the domain. Both the velocity and pressure are determined by the steady-state Navier--Stokes equation. \cite{tomasetto2025reduced} uses $\nu = 1$ and no-slip boundary conditions on the external walls and Dirichlet boundary conditions on the cylinders. They then consider a quantity (e.g., mass or particle density) $c: \Theta \times [0,T] \to \mathbb{R}$ that spreads inside the domain according to 
\begin{align}
    \partial_t c + \nabla \cdot (-\eta \nabla c + v(\boldsymbol{\mu}) c) = 0,
\end{align}
with homogeneous Neumann boundary conditions, $\eta=0.001$, $T=3$ and $v(\boldsymbol{\mu})$ as the solution of the Navier-Stokes equation given the velocity of the cylinders $\boldsymbol{\mu}$. For a fixed initial condition 
\begin{align}
    c(x, 0) = \frac{10}{\pi} \exp(-10 x_1^2 - 10 x_2^2)
\end{align}
the distribution $c$ depends directly on the fluid velocity $v(\boldsymbol{\mu})$. We collect roll-outs of $c_t$ for 400/50/50 varying $\boldsymbol{\mu}$ as our training/validation/test sets and aim to learn this distribution (independent of $t$). The relative L2 error for varying number of sensors is reported in Fig~\ref{fig:F4_pinball} and Tab~\ref{tab:physense_pinball}, example reconstructions at $m{=}10$ are shown in Fig.~\ref{fig:pinball_recon_k10}.

\paragraph{PhySense comparison.}
We retrain PhySense's Transolver mesh model on the same 400 training trajectories with their released configuration (sensor counts drawn per training step from $m \in [10,200]$), evaluated at checkpoint 78 of a planned 801 epochs. Reconstruction uses their sampler (5-step Euler) on 100 held-out snapshots. The walk stage follows their release (backbone frozen, coordinates optimised by Adam for 21 epochs at their default learning rate, their $k$-nearest-neighbour inverse-distance interpolation with $k{=}8$), run from both SSPOR (shown in Fig.~\ref{fig:F4_pinball}) and random initialisations.

\begin{table}[h]
\centering
\small
\setlength{\tabcolsep}{4pt}
\caption{\textbf{Pinball}: relative $L^2$ error of GRIFDIR reconstruction for each placement plus PhySense model (Transolver backbone).}
\label{tab:physense_pinball}
\begin{tabular}{lrrrrrrr}
\toprule
placement & $m{=}1$ & $m{=}5$ & $m{=}10$ & $m{=}15$ & $m{=}20$ & $m{=}25$ & $m{=}30$ \\
\midrule
Random & 0.961 & 0.856 & 0.741 & 0.573 & 0.496 & 0.348 & 0.197 \\
A-optimal & 0.849 & 0.555 & 0.318 & 0.181 & 0.126 & 0.077 & 0.060 \\
D-optimal & 0.849 & 0.564 & 0.346 & 0.131 & 0.079 & 0.071 & 0.048 \\
E-optimal & 0.849 & 0.511 & 0.296 & 0.192 & 0.137 & 0.129 & 0.076 \\
SSPOR & 0.877 & 0.649 & 0.471 & 0.376 & 0.224 & 0.136 & 0.110 \\
Christoffel-DPS & 0.895 & 0.723 & 0.542 & 0.454 & 0.262 & 0.158 & 0.129 \\
greedy Christoffel-DPS & 0.679 & 0.465 & 0.203 & 0.070 & 0.079 & 0.049 & 0.070 \\
ensemble Christoffel-DPS & 0.885 & 0.329 & 0.152 & 0.088 & 0.043 & 0.049 & 0.037 \\
\midrule
PhySense walk (SSPOR init) & 0.931 & 0.807 & 0.613 & 0.581 & 0.511 & 0.480 & 0.456 \\
PhySense walk (random init) & 0.850 & 0.897 & 0.762 & 0.527 & 0.428 & 0.395 & 0.433 \\
PhySense greedy Christoffel & 0.945 & 0.702 & 0.515 & 0.402 & 0.374 & 0.338 & 0.331 \\
\bottomrule
\end{tabular}
\end{table}

\begin{figure}[tbp]
  \centering
  \includegraphics[width=\linewidth]{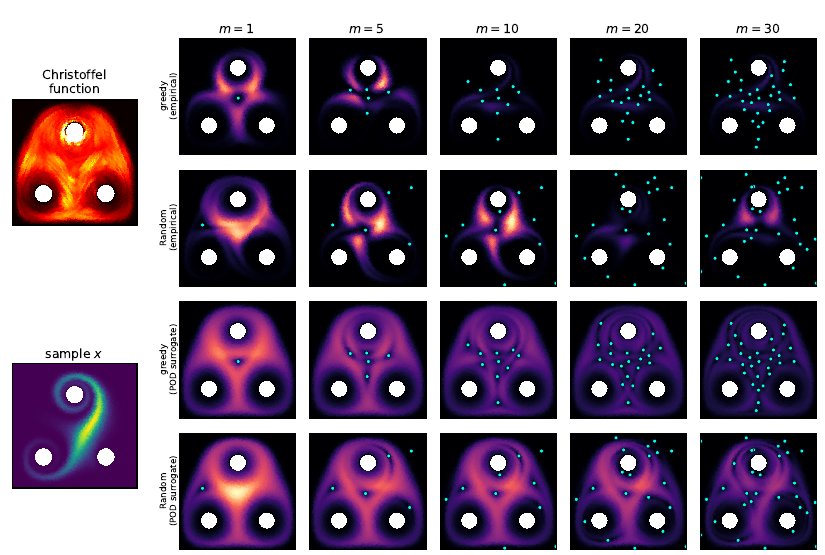}
  \caption{\textbf{Posterior collapse on Pinball.} Left: empirical Christoffel function and a training sample. Right: posterior variance given $m$ sensors, empirical (rows 1--2) and POD-OED surrogate (rows 3--4), greedy Christoffel-DPS vs random; one color scale per posterior type.}
  \label{fig:scorefield}
\end{figure}

\paragraph{Score field and posterior collapse.}
Fig.~\ref{fig:scorefield} plots the empirical Christoffel function 
$\widehat{K}_M(j) = \max_{x \in \mathbb{S}_M} |x_j|^2$ from \eqref{K-emp}, computed on $M=2000$ Pinball training snapshots, alongside a training sample, and shows how greedy Christoffel-DPS first selects symmetric points that subdivide the eight octants of the cylinders' angular rotation, maximising identifiability so the posterior collapses within a handful of sensors. We compute the posterior in two ways. Rows 1--2 are sample-based, with no Gaussian or low-rank assumption: holding out one trajectory's snapshot as ground truth, we plot the node-wise variance of the 50 training snapshots closest to its readings at the $m$ sensors. Rows 3--4 use the Gaussian POD-OED surrogate on which the baselines operate (rank-200 expansion with singular-value weights as covariance, noiseless point observations). Under both, the greedy posterior collapses by $m \approx 10$, while under random placement it remains broad at every budget.

\subsection{Darcy flow problem}\label{app:experiments-darcy}

\begin{figure}[tbp]
  \centering
  \includegraphics[width=\linewidth]{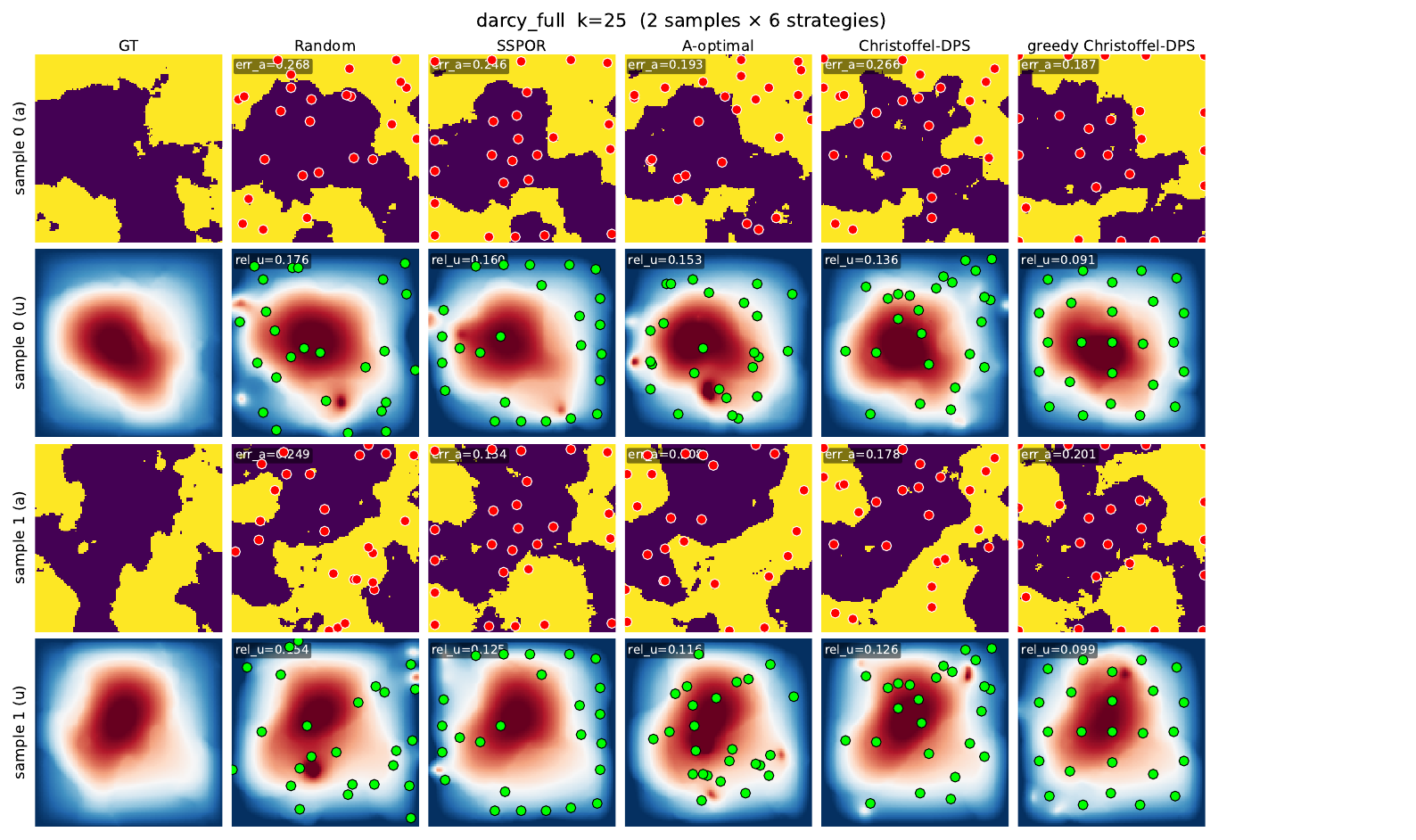}
  \caption{\textbf{Darcy reconstructions at $m{=}25$:} Two test samples; rows alternate the permeability $a$ and pressure $u$ channels. Per-cell error annotated.}
  \label{fig:darcy_recon_k25}
\end{figure}

We use the Darcy benchmark of \cite{huang_diffusionpde_2024}: the steady-state PDE
\begin{align}
  -\nabla \cdot (a \nabla u) = f \qquad \text{on } [0,1]^2,
\end{align}
with fixed source $f$, no-flux Dirichlet boundary conditions, and a binary permeability field $a \in \{3, 12\}$ piecewise-constant on a Voronoi tessellation. The pressure $u$ is the diffusion-smoothed solution. We use the official $128 \times 128$ test set (10{,}000 fields) and the pretrained EDM denoiser from \cite{huang_diffusionpde_2024}; reconstruction follows their DPS posterior-sampling variant with $500$ Heun steps and the published $\zeta$ schedule. Per-strategy reconstructions appear in Fig.~\ref{fig:darcy_recon_k25}.

\paragraph{PhySense comparison.}
We retrain PhySense's DiT grid model on the DiffusionPDE training shard (with their field normalisations, sensor counts drawn per step from $m \in [25,400]$) for 749 epochs, evaluating checkpoint 700 with their sampler (10-step Euler) on 100 held-out fields; the metric is relative $L^2$ on $u$. The walk stage uses their release (frozen backbone, 21 epochs, bilinear interpolation at continuous sensor positions), from both SSPOR (Fig.~\ref{fig:F7_darcy}) and random initialisations. Full results in Table~\ref{tab:physense_darcy}.

\begin{table}[h]
\centering
\small
\setlength{\tabcolsep}{4pt}
\caption{\textbf{Darcy}: relative $L^2$ error of reconstruction for each placement plus PhySense model (DiT backbone).}
\label{tab:physense_darcy}
\begin{tabular}{lrrrrrrrr}
\toprule
placement & $m{=}1$ & $m{=}5$ & $m{=}10$ & $m{=}25$ & $m{=}50$ & $m{=}100$ & $m{=}200$ & $m{=}500$ \\
\midrule
Random & 0.286 & 0.229 & 0.174 & 0.136 & 0.107 & 0.104 & 0.069 & 0.049 \\
A-optimal & 0.290 & 0.219 & 0.182 & 0.126 & 0.108 & 0.099 & 0.089 & 0.080 \\
D-optimal & 0.306 & 0.358 & 0.378 & 0.396 & 0.379 & 0.397 & 0.351 & 0.337 \\
E-optimal & 0.294 & 0.345 & 0.418 & 0.450 & 0.467 & 0.407 & 0.365 & 0.354 \\
D-optimal (reg.) & 0.316 & 0.305 & 0.302 & 0.338 & 0.349 & 0.392 & 0.311 & 0.330 \\
E-optimal (reg.) & 0.300 & 0.321 & 0.346 & 0.498 & 0.443 & 0.364 & 0.370 & 0.292 \\
SSPOR & 0.317 & 0.269 & 0.213 & 0.137 & 0.113 & 0.088 & 0.071 & 0.051 \\
Christoffel-DPS & 0.291 & 0.226 & 0.188 & 0.141 & 0.138 & 0.088 & 0.082 & 0.077 \\
greedy Christoffel-DPS & 0.296 & 0.198 & 0.156 & 0.108 & 0.108 & 0.062 & 0.047 & 0.032 \\
ensemble Christoffel-DPS & 0.341 & 0.314 & 0.228 & 0.144 & 0.161 & 0.097 & 0.096 & 0.066 \\
\midrule
PhySense walk (SSPOR init) & 0.644 & 0.197 & 0.126 & 0.055 & 0.036 & 0.030 & 0.025 & 0.023 \\
PhySense walk (random init) & 0.595 & 0.248 & 0.117 & 0.060 & 0.039 & 0.031 & 0.025 & 0.023 \\
PhySense greedy Christoffel & 0.664 & 0.163 & 0.096 & 0.047 & 0.035 & 0.028 & 0.024 & 0.023 \\
\bottomrule
\end{tabular}
\end{table}

\subsection{Kolmogorov flow problem}\label{app:experiments-kmflow}

\begin{figure}[tbp]
  \centering
  \includegraphics[width=\linewidth]{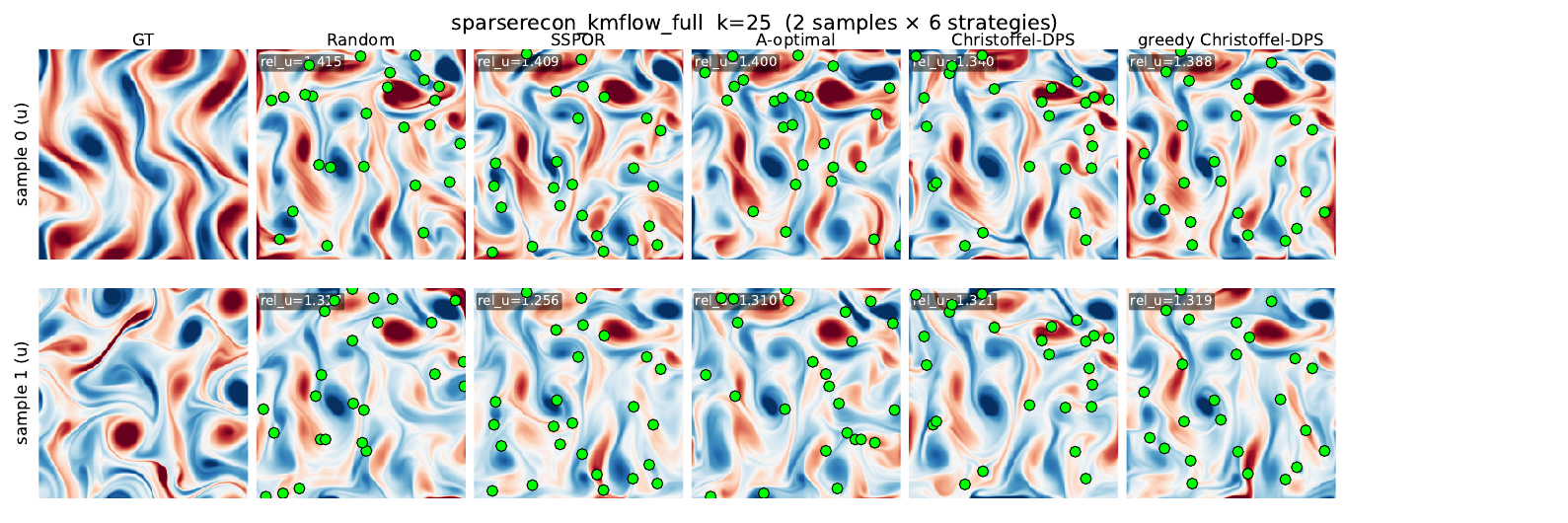}
  \caption{\textbf{Kolmogorov-flow reconstructions at $m{=}25$:} Mid-trajectory vorticity slice on the $256{\times}256$ grid. Two test samples; columns mirror Fig.~\ref{fig:darcy_recon_k25}.}
  \label{fig:kmflow_recon_k25}
\end{figure}

We use the 2D Kolmogorov turbulence benchmark of \cite{amoros-trepat_guiding_2026}: incompressible Navier--Stokes at $\mathrm{Re}{=}1000$ with sinusoidal forcing $\mathbf{f} = (\sin(8 y), 0)$ on a periodic $256 \times 256$ domain, simulated to a stationary regime. Each sample is a three-step temporal stack of vorticity. We use their pretrained DDIM checkpoint with the masked-blending sparse-reconstruction sampler ($200$ reverse-diffusion steps). Per-strategy reconstructions appear in Fig.~\ref{fig:kmflow_recon_k25}.

\begin{figure}
  \centering
  \includegraphics[width=0.75\linewidth]{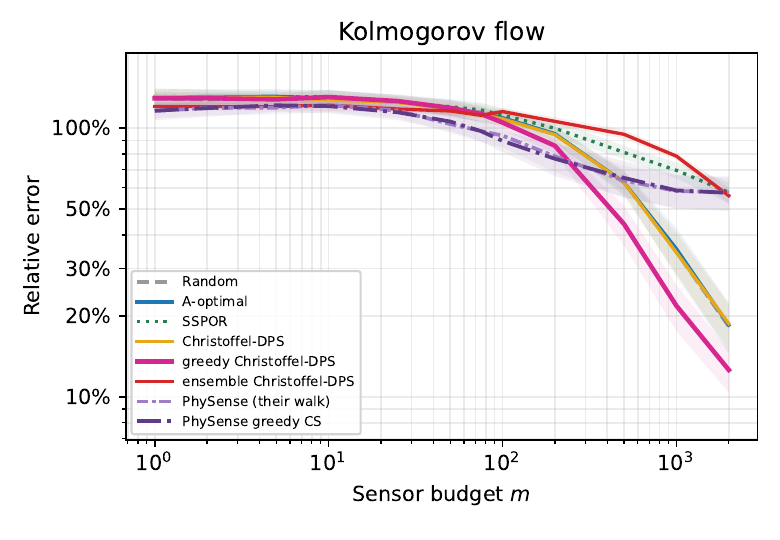}
  \caption{\textbf{Kolmogorov flow: $m$-convergence.} Relative $L^2$ error vs sensor budget $m$; bands one std. Purple: PhySense comparison (Sec.~\ref{app:experiments-kmflow}).}
  \label{fig:F_kmflow}
\end{figure}

\paragraph{PhySense comparison.}
We retrain PhySense's DiT grid model on the leading 3{,}180 vorticity stacks (their normalisation, sensor counts drawn per step from $m \in [25,400]$) for 231 epochs, evaluating checkpoint 200 on 100 held-out stacks with their 10-step Euler sampler; full results in Table~\ref{tab:physense_kmflow}. The walk follows their release from SSPOR (Fig.~\ref{fig:F_kmflow}) and random initialisations; note $m \in \{1000, 2000\}$ lies outside the trained sensor-count range for both the walk and the fixed placements.

\begin{table}[h]
\centering
\small
\setlength{\tabcolsep}{4pt}
\caption{\textbf{Kolmogorov flow}: relative $L^2$ error of reconstruction for each placement plus PhySense model (DiT backbone).}
\label{tab:physense_kmflow}
\begin{tabular}{lrrrrrrr}
\toprule
placement & $m{=}1$ & $m{=}25$ & $m{=}50$ & $m{=}100$ & $m{=}500$ & $m{=}1000$ & $m{=}2000$ \\
\midrule
Random & 1.285 & 1.248 & 1.190 & 1.094 & 0.629 & 0.345 & 0.182 \\
A-optimal & 1.307 & 1.255 & 1.189 & 1.092 & 0.627 & 0.355 & 0.185 \\
SSPOR & 1.310 & 1.250 & 1.201 & 1.117 & 0.812 & 0.696 & 0.576 \\
Christoffel-DPS & 1.309 & 1.244 & 1.180 & 1.082 & 0.627 & 0.344 & 0.187 \\
greedy Christoffel-DPS & 1.290 & 1.260 & 1.189 & 1.049 & 0.439 & 0.218 & 0.126 \\
ensemble Christoffel-DPS & 1.203 & 1.177 & 1.158 & 1.152 & 0.948 & 0.786 & 0.560 \\
\midrule
PhySense walk (SSPOR init) & 1.158 & 1.161 & 1.035 & 0.938 & 0.637 & 0.583 & 0.576 \\
PhySense walk (random init) & 1.149 & 1.156 & 1.058 & 0.947 & 0.635 & 0.587 & 0.579 \\
PhySense greedy Christoffel & 1.161 & 1.144 & 1.058 & 0.896 & 0.652 & 0.586 & 0.574 \\
\bottomrule
\end{tabular}
\end{table}

\section{Christoffel-DPS algorithms}
\label{app:algorithms}

This appendix expands the offline and online algorithms summarized in
\S\ref{s:cdps} into explicit pseudocode.  All algorithms are stated in
the finite grid setting of \S\ref{s:cdps}. The unknown signal $f^*$ is identified with $x^*\in\bbR^N$, sensors with rows of
$I_N$ collected into the row-selector matrix $S$ and measurements with
$y_S = Sx^* + \eta$.

\subsection{Offline Christoffel-DPS}
\label{app:offline}

The greedy variant requires further discussion, specifically, the implementation via QR with column-pivoting. Algorithm~\ref{alg:offline-fingerprint} greedily computes the rows of $S$ by greedy Gram--Schmidt deflation on the matrix $X$ from \S \ref{ss:offline}.  At each iteration, the
location with the largest residual norm is selected and the
value corresponding every remaining location is deflated along the just-picked
direction. This is equivalent to column-pivoted QR
\citep{businger_linear_1965} on $X$, which scipy implements as
\verb|scipy.linalg.qr(X.T, pivoting=True)|; we state the deflation
form for better readability and understanding.

\noindent\textbf{Complexity.}  Each iteration involves $O(N M)$ flops for the
deflation and $O(N)$ for the argmax, giving a total of $O(m N M)$ flops.  Householder
pivoted QR has the same scaling.

\begin{algorithm}[H]
\caption{Greedy Christoffel-DPS}
\label{alg:offline-fingerprint}
\begin{algorithmic}[1]
\Require mean-adjusted snapshot matrix $X\in\bbR^{N\times M}$, sensor budget $m$
\Ensure sensor index sequence $(s_1,\ldots,s_m)$
\State $R \gets X$ \Comment{initialize residual, $R_j$ is row $j$ of $R$}
\For{$i = 1,\ldots,m$}
  \State $s_i \gets \arg\max\limits_{j \notin \{s_1,\ldots,s_{i-1}\}}
                              \nm{R_j}^2$
         \Comment{pixel with largest residual norm}
  \State $q_i \gets R_{s_i} / \nm{R_{s_i}}$
  \For{$j = 1,\ldots,N$}
    \State $R_j \gets R_j - \langle R_j, q_i\rangle\,q_i$
           \Comment{deflate along $q_i$}
  \EndFor
\EndFor
\State \Return $(s_1,\ldots,s_m)$
\end{algorithmic}
\end{algorithm}

\subsection{Online ensemble Christoffel-DPS}
\label{app:online}

Algorithm~\ref{alg:online-cdps} runs an ensemble of $N_e$ DPS chains
through a standard reverse-diffusion schedule.  The $m$ sensors are
split into $m_0$ \emph{anchor} sensors at fixed positions typically chosen by the offline greedy Christoffel ordering of \S\ref{ss:offline} and $m_1 = m - m_0$ \emph{mobile} sensors that drift through the domain at a sequence of \emph{drift events}.  At a drift event the empirical Christoffel function~\eqref{K-online} is evaluated on the live ensemble's Tweedie point estimates and each mobile sensor moves to the highest-scoring unoccupied node within a fixed-radius $r_{\rm drift}$-neighbourhood of its current position; the per-event displacement is bounded by $r_{\rm drift}$, so the total mobile-sensor travel over a reverse pass is at most $D \cdot r_{\rm drift}$.  Chains whose measurement residual is too large are pruned at the same events.

For convenience, we write $\textsc{ReverseStep}$ for one step of the underlying diffusion sampler a Heun or Euler predictor-corrector update of the form
$z^{(i)}_{t_{k+1}} = \mathrm{Pred}(z^{(i)}_{t_k}, \hat x^{(i)}_0;\sigma(t_k),\sigma(t_{k+1}),C)
- \alpha_k\,g^{(i)}$ \citep{karras_elucidating_2022,chung2023diffusion},
where $\alpha_k$ is the DPS guidance weight and $g^{(i)}$ is the measurement gradient defined at line~7 of Algorithm~\ref{alg:online-cdps}.

\noindent\textbf{Hyperparameters used in the experiments.}  $N_e = 20$
chains and a logarithmic drift schedule with $D = 20$ events on
DiffusionPDE benchmarks (Algorithm~\ref{alg:online-cdps} with
$\Delta\ell = \infty$, i.e.\ no pruning) and $D = 10$ events on the
Pinball benchmark with $\Delta\ell = 1$, $N_{\min} = 1$, and $m_0 = 3$
anchor sensors taken from the offline empirical-Christoffel ordering.
Allocation $(n_1,\ldots,n_D)$ is uniform: $n_d = (m - m_0)/D$.
Measurement noise $\sigma_\eta = 0.1$; DPS guidance weight $\alpha_k$
inside $\textsc{ReverseStep}$ follows \citet{chung2023diffusion}
with linear schedule and unit weight at $\sigma(t) = \sigma_\eta$.

\noindent\textbf{Limits.}  Setting $D = 0$ (no drift events) reduces
Algorithm~\ref{alg:online-cdps} to a fixed-sensor DPS sampler with the
offline empirical-Christoffel selection of
Algorithm~\ref{alg:offline-fingerprint} fed in through $S$.  Replacing
the secant numerator at line~13 by the per-pixel ensemble standard
deviation $\mathrm{Var}_e(\hat x^{(e)}_{0,j})^{1/2}$ recovers the ensemble-uncertainty score of \citet{chakraborty_adaptive_2026}, a second-moment summary in place of the sup over secants. Their selection protocol differs: sensors are taken from a single score map per assimilation step with topology-aware suppression, whereas Algorithm~\ref{alg:online-cdps} relocates its mobile sensors at scheduled drift events within a single reverse pass.

\noindent\textbf{Complexity.}  Each reverse-diffusion step costs
$O(N_e \cdot \mathrm{cost}(D_\theta))$.  Each drift event costs
$O(N_e^2 |J|)$ for the pairwise score plus $O(N_e^2 m)$ for residual
evaluation, with $|J| \le N$ the unobserved budget.  Score evaluation
is typically dominated by the denoiser cost.

\begin{algorithm}[H]
\caption{Online ensemble Christoffel-DPS}
\label{alg:online-cdps}
\begin{algorithmic}[1]
\Require denoiser $D_\theta$; covariance $C$; reverse-time schedule
$T = t_0 > \cdots > t_K = 0$ with noise levels $\sigma(t_k)$;
drift schedule $\cT = \{t_{d_1},\ldots,t_{d_D}\}$; number of anchor sensors $m_0$ and drift sensors $m_1$, total number of sensors $m$; initial selector $S$ (first $m_0$ rows correspond to the anchor sensor locations);node coordinates $\xi_1,\ldots,\xi_N \in \bbR^d$; drift radius $r_{\rm drift}$; ensemble size $N_e$; pruning gap $\Delta\ell \ge 0$ with floor
$N_{\min}\ge 1$; ground-truth signal $x^*$ 

\Ensure reconstruction $\hat x \in \bbR^N$
\State $y_S \gets S x^*$
       \Comment{measurements at initial $m$ sensor locations}
\State Sample $z^{(i)}_T \sim \cN(0,\sigma^2(T)\,C)$ for $i = 1,\ldots,N_e$
\State $A \gets \{1,\ldots,N_e\}$ \Comment{surviving chain indices}
\For{$k = 0,\ldots,K-1$}
  \For{$i \in A$ \textbf{in parallel}}
    \State $\hat x^{(i)}_0 \gets D_\theta(z^{(i)}_{t_k},\sigma(t_k))$
           \Comment{Tweedie point estimate}
    \State $g^{(i)} \gets \nabla_{z^{(i)}_{t_k}}
              \tfrac{1}{2\sigma_\eta^2}\nm{S\hat x^{(i)}_0 - y_S}^2$
           \Comment{measurement gradient}
    \State $z^{(i)}_{t_{k+1}} \gets
              \textsc{ReverseStep} \bigl(z^{(i)}_{t_k},\hat x^{(i)}_0,g^{(i)};\sigma(t_k),\sigma(t_{k+1}),C\bigr)$
  \EndFor
  \If{$t_{k+1} \in \cT$ (drift event)}
   \Statex \hspace{1em}\textit{// Score, drift mobile sensors, observe, prune}
    \State $J \gets \{1,\ldots,N\} \setminus \mathrm{rows}(S_{1:m_0})$
           \Comment{drift sensor candidate indices}
    \For{$j \in J$}
      \State $\widehat{K}(j) \gets
              \displaystyle\max_{l\neq l' \in A}\;
              \frac{(\hat x^{(l)}_{0,j} - \hat x^{(l')}_{0,j})^2}
                   {\nm{\hat x^{(l)}_0 - \hat x^{(l')}_0}^2}$
    \EndFor

    \For{$l = 1,\ldots,m_1$}
        \State Draw $j_l \propto \widehat{K}$ with $\|\xi_{l} - \xi_{j_l}\| \le r_{\rm drift}$ \Comment{sample drift location within radius}
    \EndFor
    
    \State Replace the last $m_1$ rows of $S$ with rows 
    $j_1,\ldots,j_{m_1}$ of $I_N$
    \State $y_S \gets S x^*$
           \Comment{read newly observed entries}
    \State $\ell_i \gets -\tfrac{1}{2\sigma_\eta^2}
              \nm{S\hat x^{(i)}_0 - y_S}^2$ for $i \in A$
           \Comment{measurement log-likelihood}
    \State $\ell^\star \gets \max_{i \in A} \ell_i$
    \State $A \gets \{ i \in A : \ell^\star - \ell_i \le \Delta\ell |S|\}$
           \Comment{prune chains}
    \If{$|A| < N_{\min}$}
      \State retain the $N_{\min}$ chains with largest $\ell_i$
             \Comment{floor on alive ensemble size}
    \EndIf
  \EndIf
\EndFor
\Statex \textit{// Collapse the ensemble}
\State $i^\star \gets \arg\max_{i\in A} \ell_i$
\State \Return $\hat x \gets \hat x^{(i^\star)}_0$
       \quad\textbf{or}\quad
       $\hat x \gets \tfrac{1}{|A|}\sum_{i\in A}\hat x^{(i)}_0$
\end{algorithmic}
\end{algorithm}

\end{document}